\documentclass[letterpaper]{article} 
\usepackage{aaai25}  
\usepackage{times}  
\usepackage{helvet}  
\usepackage{courier}  
\usepackage[hyphens]{url}  
\usepackage{graphicx} 
\usepackage{natbib}  
\usepackage{caption} 
\usepackage{algorithm}
\usepackage{amssymb}
\usepackage{afterpage}
\usepackage{subcaption}
\usepackage{xcolor}
\usepackage{algpseudocode}
\usepackage{amsmath}
\usepackage{multirow}
\usepackage{booktabs}
\usepackage{bm}

\DeclareMathOperator*{\argmax}{argmax} 
\newcommand{\x}{{\bf x}}

\newcommand{\w}{{\bm w}}

\newcommand{\D}{\mathcal{D}}
\newcommand{\Y}{\mathcal{Y}}
\newcommand{\A}{\mathcal{A}}

\newcommand{\R}{\mathbb{R}}

\newcommand{\eg}{\emph{e.g.}}
\newcommand{\ie}{\emph{i.e.}}
\newcommand{\name}{{\sc MOS }}
\newcommand{\mame}{{\sc MOS}}
\newcommand{\APER}{{\sc Aper}}

\usepackage{colortbl}
\definecolor{LightCyan}{rgb}{0.88,1,1}
\definecolor{Gray}{gray}{0.85}
\definecolor{myblue}{rgb}{0.88,0.98,1}

\usepackage{newfloat}
\usepackage{listings}
\DeclareCaptionStyle{ruled}{labelfont=normalfont,labelsep=colon,strut=off} 
\floatstyle{ruled}
\newfloat{listing}{tb}{lst}{}
\floatname{listing}{Listing}
\title{MOS: Model Surgery for \\Pre-Trained Model-Based Class-Incremental Learning}
\author{
    Hai-Long Sun\textsuperscript{\rm 1, \rm 2},
    Da-Wei Zhou\textsuperscript{\rm 1, \rm 2}\footnotemark[1],
    Hanbin Zhao\textsuperscript{\rm 3},
    Le Gan\textsuperscript{\rm 1, \rm 2},
    De-Chuan Zhan\textsuperscript{\rm 1, \rm 2},
    Han-Jia Ye\textsuperscript{\rm 1, \rm 2}\thanks{Corresponding author.}
}
\affiliations{
    \textsuperscript{\rm 1}School of Artificial Intelligence, Nanjing University\\
    \textsuperscript{\rm 2}National Key Laboratory for Novel Software Technology, Nanjing University\\
    \textsuperscript{\rm 3}College of Computer Science and Technology, Zhejiang University\\

    \{sunhl, zhoudw, zhandc, yehj\}@lamda.nju.edu.cn, ganle@nju.edu.cn, zhaohanbin@zju.edu.cn
}

\usepackage{bibentry}

\begin{document}

\maketitle

\begin{abstract}
    
    Class-Incremental Learning (CIL) requires models to continually acquire knowledge of new classes without forgetting old ones. 
    Despite Pre-trained Models (PTMs) have shown excellent performance in CIL, catastrophic forgetting still occurs as the model learns new concepts.
    Existing work seeks to utilize lightweight components to adjust the PTM, while the forgetting phenomenon still comes from {\em parameter and retrieval} levels.
    Specifically, iterative updates of the model result in parameter drift, while mistakenly retrieving irrelevant modules leads to the mismatch during inference. 
    To this end, we propose MOdel Surgery (\mame) to rescue the model from forgetting previous knowledge.
    By training task-specific adapters, we continually adjust the PTM to downstream tasks. 
    To mitigate parameter-level forgetting, we present an adapter merging approach to learn task-specific adapters, which aims to bridge the gap between different components while reserve task-specific information. 
    Besides, to address retrieval-level forgetting, we introduce a training-free self-refined adapter retrieval mechanism during inference, which leverages the model's inherent ability for better adapter retrieval.
    By jointly rectifying the model with those steps, \name can robustly resist catastrophic forgetting in the learning process.
    Extensive experiments on seven benchmark datasets validate \mame's state-of-the-art performance. Code is available at: \url{https://github.com/sun-hailong/AAAI25-MOS}

\end{abstract}

\section{Introduction}

In recent years, deep learning has achieved significant results in many real-world applications~\cite{deng2009imagenet,he2015residual,cao2024domain-p,sun2024parrot,ye2019learning,yang2023lever,zhang2024from}. While in the open world, data often appears in a streaming format, requiring a machine learning paradigm capable of incrementally acquiring new class knowledge, which is denoted as Class-Incremental Learning (CIL)~\cite{rebuffi2017icarl,zhou2024continual}. One of the significant challenges in CIL is catastrophic forgetting, where the model, after learning new classes incrementally, gradually loses its ability to recognize the old ones~\cite{french1999catastrophic}. In response to this challenge, the field of CIL is evolving with the emergence of pre-trained models (PTMs). Unlike the traditional approach of ``training from scratch''~\cite{li2017learning,zhou2022model}, contemporary CIL methods are increasingly leveraging PTMs, which are initially pre-trained on vast datasets using substantial resources~\cite{mcdonnell2024ranpac,jung2023generating}. This pre-training process endows PTMs with robust generalization abilities. Consequently, designing an effective CIL method that leverages PTMs and resists catastrophic forgetting has garnered significant attention from researchers.

Due to the generalization of PTMs, existing works often freeze the pre-trained weights and adapt to incremental tasks using additional lightweight modules~\cite{hu2022lora,rebuffi2017learning,chao2020revisiting,ye2022generalized}. For example, visual prompt tuning~\cite{jia2022visual} customizes prompts to modify model behavior, facilitating adaptation to downstream tasks. Specifically, L2P~\cite{wang2022learning} designs a key-query matching strategy to retrieve instance-specific prompts from a prompt pool. Based on L2P, DualPrompt~\cite{wang2022dualprompt} introduces expert prompts to encode task-specific information and explores the impact of prompt depth. Furthermore, CODA-Prompt~\cite{smith2023coda} proposes an attention-based weighting method for prompts to enhance the efficacy of prompt retrieval. 

However, as the model learns new concepts, catastrophic forgetting still occurs. This forgetting phenomenon happens at both the {\em parameter and retrieval levels}. During the training stage, although many methods use lightweight components to adjust the PTM, iterative updates of these components will lead to parameter drift and trigger forgetting. Moreover, existing works devote to preventing conflicts between prompts or achieving orthogonal projection, which exacerbates parameter drift between new and old components. During inference, training multiple lightweight modules requires selecting the most relevant one, but the model may mistakenly retrieve the irrelevant modules, leading to the performance decay. This motivates us to question if it is possible to {\em jointly rectify the model to resist catastrophic forgetting at both the parameter and retrieval levels?}

Facing the challenges at both the parameter and retrieval levels, our model should be able to effectively design mechanisms to overcome these issues. To address forgetting at the parameter level, the model needs to develop effective update methods that ensure the updated parameters remain discriminative for old data. To overcome forgetting at the retrieval level, the model requires efficient self-correction strategies to help utilize relevant information, assisting in the instance-specific retrieval of lightweight modules. \looseness=-1

To this end, we propose MOdel Surgery (\mame) for pre-trained model-based class-incremental learning to rescue the model from forgetting previous knowledge. This surgery is divided into the training and inference stages. To mitigate parameter-level forgetting, we present an {\em adapter merging} approach during training, which learns task-specific adapters while bridging gaps between components and retaining task-specific information. This strategy helps previously learned adapters aid in learning new tasks. To address retrieval-level forgetting, we introduce a training-free \emph{self-refined adapter retrieval mechanism} during inference, which leverages the model's inherent ability for better adapter retrieval. This mechanism requires no additional training overhead, making the algorithm simple and efficient.
Finally, to enable the model to balance the stability-plasticity dilemma, we present a model ensemble method that integrates the model's capabilities across multiple phases. It not only ensures strong generalization but also allows the model to quickly recognize and update information. 
Experiments on seven benchmark datasets validate the effectiveness of \mame. Additionally, the visualization of the self-refined adapter retrieval mechanism indicates that \name effectively learns adapter retrieval for various downstream tasks. \looseness=-1
\section{Related Work}
\noindent\textbf{Class-Incremental Learning (CIL).} It aims to enable models to acquire new classes knowledge while retaining previously learned information~\cite{rebuffi2017icarl}. Existing works can be roughly categorized into several categories. Knowledge distillation-based methods~\cite{li2017learning,rebuffi2017icarl,snell2017prototypical} establish a mapping between the former stage model and the current model, thereby aiding the latter in retaining characteristics from earlier updates during incremental learning~\cite{hinton2015distilling}. Data rehearsal-based methods~\cite{chaudhry2018efficient,liu2020mnemonics,zhao2021memory} select and replay crucial exemplars from old classes during training new ones to continuously revise former knowledge. Parameter regularization-based methods~\cite{aljundi2019task,kirkpatrick2017overcoming} aim to predict and minimize the drift of key parameters by using regularization terms. Model rectification-based methods~\cite{pham2021continual,shi2022mimicking,yu2020semantic} focus on correcting the model's inductive bias to ensure unbiased estimations. Model expansion-based methods~\cite{chen2023dynamic,hu2023dense,wang2022foster,yan2021dynamically} construct non-interfering subnetworks for each task. During inference, they are combined to form a larger feature map and train a classifier to effectively calibrate across all classes.

\noindent\textbf{Pre-Trained Model-Based CIL.} PTM-based CIL has emerged as a hot topic in the current CIL research area. With advances in pre-training techniques, numerous parameter-efficient fine-tuning (PEFT) methods~\cite{jia2022visual,hu2022lora,lian2022scaling,rebuffi2017learning,cao2024domain-c,hutask,lu2024visual,li2024configure,wei2019multi,zhang2024seeing} have been developed. These methods aim to improve model performance with minimal additional resources while freezing pre-trained weights. In this context, L2P~\cite{wang2022learning} introduces a prompt pool, selecting instance-specific prompts via a key-query matching selection mechanism to guide the PTM's response. DualPrompt~\cite{wang2022dualprompt} extends L2P by designing G-Prompt and E-Prompt, which encode task-invariant and task-specific instructions, respectively. CODA-Prompt~\cite{smith2023coda} innovates by developing decomposed prompts and combining them using an attention-based weighting method. DAP~\cite{jung2023generating} extends prompt selection into prompt generation. SLCA~\cite{zhang2023slca} reveals that fine-tuning a ViT backbone with a lower learning rate at the representation layer yields higher accuracy than prompt strategies. APER~\cite{zhou2024revisiting} explores various PEFT methods and shows that prototypical classifiers serve as a strong baseline, and RanPAC~\cite{mcdonnell2024ranpac} further expands APER in random projection. EASE~\cite{zhou2024expandable} concatenates the feature
representations of multiple task-specific backbones.
\section{Preliminaries}
\subsection{Class-Incremental Learning}

Class-incremental learning aims to acquire knowledge from continuously evolving data streams that introduce new classes while retaining knowledge of previous ones to build a unified classifier~\cite{rebuffi2017icarl}. Consider a series of $B$ training stages, expressed as $\{\D^1,\D^2,\cdots,\D^B\}$, where $\D^b=\{(\x_{i}^{b}, y_{i}^{b})\}_{i=1}^{n_b}$ represents the $b$-th incremental stage containing $n_b$ instances. Correspondingly, the testing set is denoted as $\{\D_t^1,\D_t^2,\cdots,\D_t^B\}$. Within this setting, each training instance $\x_{i}^{b} \in \R^{D}$ is associated with a class $y_i \in Y_b$. Here, $Y_b$ defines the set of labels for task $b$, and it is ensured that $Y_b \cap Y_{b^{\prime}}=\varnothing$ for any $b\neq b^\prime$. During $b$-th training stage, the model is updated utilizing data exclusively from $\D^b$. In this paper, we follow the \textbf{exemplar-free setting} in~\cite{wang2022learning,wang2022dualprompt,zhou2024revisiting}, which entails not using any historical exemplars from previous classes. Therefore, the model can only access data from $\D^b$ for training during the $b$-th stage. The effectiveness of the model is evaluated across all previously encountered classes, collectively represented as $\mathcal{Y}_b = Y_1 \cup \cdots \cup Y_b$, after each CIL task. Specifically, we aim to find a model $f(\x): X \rightarrow \mathcal{Y}_b$ that minimizes empirical risk across all test datasets:
\begin{equation} \label{eq:cilrisk} 
f^*=\underset{f\in\mathcal{H}}{\operatorname*{argmin}} \ \mathbb{E}_{(\mathbf{x},y)\sim\D_t^1\cup\cdots\D_t^b}\mathbb{I}\left(y\neq f(\mathbf{x})\right),
\end{equation}
where $\mathcal{H}$ is the hypothesis space and $\mathbb{I}(\cdot)$ denotes the indicator function. $\D_t^b$ represents the testing set of task $b$. An effective CIL model satisfying Eq.~\ref{eq:cilrisk} exhibits discriminative abilities across all classes. It achieves a balance between learning new classes and retaining information about old ones. \looseness=-1

Following the typical PTM-based CIL works~\cite{wang2022learning,wang2022dualprompt}, we assume that a PTM (\eg, Vision Transformer (ViT)~\cite{dosovitskiy2020image}) is available as the initialization for $f(\x)$. For clearer understanding, we decouple the PTM into two components: $f(\x)=W^{\top}\phi(\x)$, where $\phi(\cdot):\mathbb{R}^{D} \rightarrow \mathbb{R}^{d}$ is the feature extractor and $W\in\mathbb{R}^{d\times |\mathcal{Y}_{b}|}$ is the classifier. We denote the classifier for class $k$ as $\w_k$: $W=[\w_1,\w_2,\cdots,\w_{|\Y_{b}|}]$. For a standard ViT, the initial encoding layer converts the image into a sequence of output features, denoted as $\x_e \in \R^{L \times d}$, where $L$ is the sequence length. We simplify this by treating the first token in $\x_e$ to be the \texttt{[CLS]} token. The sequence $\x_e$ is then processed through subsequent layers, including multi-head self-attention and MLP, to produce the final embeddings. Finally, the embedded \texttt{[CLS]} token is considered as $\phi(\x)$.

\begin{figure*}[t]
	\begin{center}
		\includegraphics[width=1.8\columnwidth]{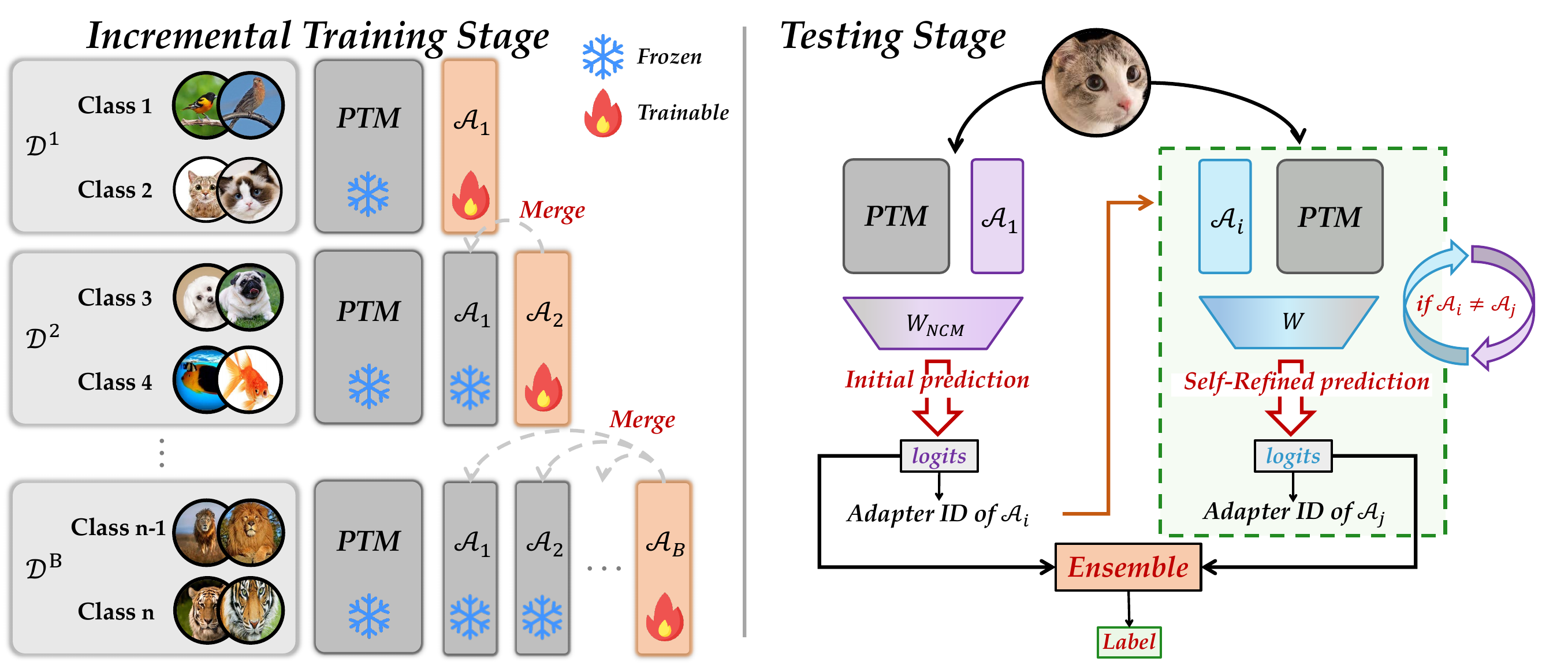}
	\end{center}
	\caption{\small  Illustration of \mame. {\bf Left}: the training protocol of \mame. We use progressively merged adapters to incrementally adapt the PTM.
	{\bf Right}: the self-refined adapter retrieval mechanism for the testing stage. We use the model's own capabilities to correct errors caused by the mistaken retrieval problem.
	} 
    \label{figure:teaser}
\end{figure*} 

\subsection{Analysis of PTM-Based CIL}
\noindent\textbf{Learning with PTMs.}
A representative work in class-incremental learning using PTMs is the L2P~\cite{wang2022learning} approach. They introduce a strategy of freezing the pre-trained weights and constructing a learnable prompt pool that can be shared across all tasks. This prompt pool is denoted as $\mathbf{P}=\{P_1,P_2,\cdots,P_M\}$, where $P_j\in\mathbb{R}^{L_p\times d}$ is a single prompt with token length $L_p$ and the same embedding size $d$ as $\x_e$. $M$ is the size of the prompt pool. Each prompt in this pool corresponds to a specific key $\{(\boldsymbol{k}_1,P_1),(\boldsymbol{k}_2,P_2),\cdots,(\boldsymbol{k}_M,P_M)\}$, where $\boldsymbol{k}_i\in\mathbb{R}^{d_k}$. First, they utilize a PTM without prompting (\ie, $\phi(\cdot)$) to encode the features into the key's embedding space and retrieve prompts with similar keys.  During inference, given an input $\x$, the model employs $\phi(\x)$ to look up the top-N keys by solving the objective in Eq.~\ref{eq:key-value}. This process retrieves the most relevant keys and their corresponding prompts from the prompt pool. \looseness=-1
\begin{equation} \label{eq:key-value} 
    \mathbf{K}_{\x}=\underset{\{s_i\}_{i=1}^N\subseteq[1,M]}{\operatorname*{argmin}}\sum_{i=1}^N\gamma\left(\phi(\x),\boldsymbol{k}_{s_i}\right),
\end{equation}
where $\mathbf{K}$ is the set of all keys and $\mathbf{K}_{\x}$ is the selected top-N keys. $\gamma(\cdot, \cdot)$ denotes the cosine distance. Finally, L2P minimize the end-to-end training loss function:
\begin{equation} \label{eq:pool-opt} \min_{\mathbf{P},\mathbf{K},\phi}\ell(W^{\top}\phi(\x;\mathbf{P}),y)+\lambda\sum_{\mathbf{K}_\x}\gamma\left(\phi(\x),\boldsymbol{k}_{s_i}\right),
\end{equation}
where $\ell(\cdot,\cdot)$ is the cross-entropy loss that measures the discrepancy between prediction and ground truth. $\lambda$ is a scalar to weight the loss. Optimizing Eq.~\ref{eq:pool-opt} enhances the PTM's ability to incorporate task-specific information, allowing it to adapt more effectively to evolving data instances.

\noindent\textbf{Forgetting of parameter and retrieval levels.} 
L2P continually updates prompts and retrieves instance-specific prompts to guide the PTM's response. However, although the model learns new concepts, catastrophic forgetting still occurs at the {\em parameter and retrieval} levels. 
Specifically, Eq.~\ref{eq:pool-opt} shows how L2P uses lightweight modules to adjust the PTM to downstream tasks. As the prompts are iteratively updated, they gradually adapt to the subsequent tasks, leading to parameter drift.
On the other hand, training multiple lightweight modules requires selecting the most relevant one during inference, while the model may mistakenly retrieve the irrelevant modules, leading to the performance decay. 
The mistaken retrieval comes from three aspects: First, modules learned in previous tasks might be re-selected for new tasks, causing confusion between the retrieval of old and new modules. Besides, since the keys for subsequent tasks do not exist during current training, a gap may arise between the keys and the feature embeddings, leading to mistaken retrieval during inference. 
Therefore, it is essential to design a method to jointly rectify the model to resist catastrophic forgetting at both the parameter and retrieval levels.
\section{{\scshape{MOS}}: Model Surgery for PTM-based CIL}
Facing the challenge of resisting catastrophic forgetting, we need a method to jointly rectify the model. The key idea of \name is to design {\em model surgery} in two aspects, \ie, training stage surgery that mitigates parameter drift and testing stage surgery that retrieves better lightweight modules. Training stage surgery aims to use previously learned knowledge to improve performance on current tasks, allowing the model to adapt to new tasks more quickly. Testing stage surgery seeks to find a mechanism for better adapter retrieval without additional overhead. As a result, the model can benefit from continual lightweight module updates and effective retrieval ability without forgetting existing knowledge. \looseness=-1

We first introduce the process of progressively merged adapters for mitigating parameter drift and then discuss the self-refined adapter retrieval mechanism. We summarize the inference function with pseudo-code in the last part.

\subsection{Progressively Merged Adapters} \label{sec:merge}
To handle the parameter drift caused by the iterative updates of the model, we need to bridge the gap between different lightweight modules. In other words, as the model continually receives new data and tasks, it is crucial to effectively retain and utilize previously learned knowledge. This approach allows the model to transfer prior knowledge to new tasks and mitigates the parameter drift problem. In Eq.~\ref{eq:pool-opt}, the embedding of a given input $\x$ is obtained using instance-specific prompts. During the incremental phase, a potential problem can emerge, \ie, iterative updates to existing prompts might cause them to better match new tasks, possibly resulting in forgetting older tasks.\looseness=-1

Due to the large prompt pool in the above methods, which exacerbates mistaken retrieval, we suggest mitigating this problem by using a smaller number of lightweight modules. In detail, by directly incorporating adapter tuning~\cite{rebuffi2017learning} into the PTM to optimize a single adapter for encoding task-specific information, we achieve this goal through the application of this method. This enhanced integration allows facilitates a more effective assimilation of task-specific information. With this approach, we only need to optimize a collection of adapters to encode task-specific information.
Denote that there are $L$ transformer blocks in the pre-trained model, each with a self-attention module and an MLP layer. We integrate an adapter into each layer's MLP via residual connections. An adapter is a bottleneck module comprising a down-projection layer $W_{down} \in \R^{d\times r}$, a non-linear activation function ReLU, and an up-projection layer $W_{up} \in \R^{d\times r}$. The output formula of the MLP is formatted as follows:
\begin{align} \label{eq:adapter-define}
	\x_o = \text{MLP}(\x_i)+\text{ReLU}(\x_i W_{down})W_{up} ,
\end{align}
where $\x_i$ and $\x_o$ are the input and output of the MLP, respectively. Eq.~\ref{eq:adapter-define} illustrates how to enhance the task information by adding residual connections of adapters to the original outputs. In the context of ViT and for a specific $i$-th task, we define the set of adapters across all $L$ transformer blocks as $\mathcal{A}_i$, representing task-specific adapters. Furthermore, we denote the output embedding of a given $\mathcal{A}_i$, combined with the PTM, as $\phi(\x;\mathcal{A}_i)$. Therefore, when a new task emerges, we freeze the weights of the PTM and focus solely on optimizing the adapters and the corresponding classifier $W$:
\begin{align} \label{eq:adapter-opt}
    \min_{\mathcal{A}_i, W}\sum_{(\x,y)\in\mathcal{D}^b}\ell\left(W^\top{\phi}\left(\x;\mathcal{A}_i\right),y\right).
\end{align}

We enable the incorporation of task-specific information into embeddings through adapters by optimizing Eq.~\ref{eq:adapter-opt}, facilitating the learning of new tasks.
In an ideal scenario, if the task ID of each test sample is known, we can easily select the corresponding task-specific adapter using this ID to achieve optimal results.

However, in the CIL setting, obtaining such a task ID during the testing phase is forbidden. To address this challenge and mitigate parameter drift, we propose the training stage surgery which uses adapter merging strategy based on Exponential Moving Average (EMA) in Eq.~\ref{eq:adapter-merge}. This approach allows subsequent adapters to retain some knowledge of their predecessors, ensuring satisfactory results even if an incorrect $\mathcal{A}$ is selected.
\begin{align} \label{eq:adapter-merge}
    {\mathcal{A}_b}=(1-\alpha)\hat{\mathcal{A}_b}+\frac{\alpha}{b-1}\sum\nolimits_{k=1}^{b-1}\mathcal{A}_k ,
\end{align}
where $\hat{\mathcal{A}_b}$ represents the set of adapters for the $b$-th training stage and $\mathcal{A}_b$ is the final result after the EMA process. Specifically, given an adapter comprises $W_{up}$ and $W_{down}$, we perform the merge process on both of them to facilitate the integration of adapters. When training a new $\mathcal{A}_b$, all previously trained $\mathcal{A}_k$ are frozen, and the adapter merging process is executed following each backpropagation step. 

\noindent\textbf{Effect of adapter merging strategy.} Figure~\ref{figure:teaser} (left) depicts this merging process. This strategy ensures that the training of the current adapter $\mathcal{A}_b$ does not interfere with the performance of already trained adapters, thereby preventing catastrophic forgetting. Moreover, it guarantees that each $\mathcal{A}$ retains task-specific information while maintaining remaining well-aligned in the feature space, even if an incorrect $\mathcal{A}$ is selected. In this way, we can mitigate parameter drift during iterative adapter updates. Moreover, because adapters are lightweight branches, they require significantly fewer parameters compared to fully fine-tuning the backbone. The parameter cost for saving these adapters is calculated as $(B \times L \times 2dr)$, where $B$ denotes the number of tasks, $L$ is the number of transformer blocks, and $2dr$ signifies the parameter count of each adapter (\ie, linear projections).

\begin{table*}[t]
	\caption{\small Average and last performance comparison on seven datasets with {\bf ViT-B/16-IN21K} as the backbone.  `IN-R/A' stands for `ImageNet-R/A,' `ObjNet' stands for `ObjectNet,' and `OmniBench' stands for `OmniBenchmark.' 
		We report all compared methods with their source code.
		The best performance is shown in bold. All methods are implemented without using exemplars.
	}\label{tab:benchmark}
	\centering
	\resizebox{1.0\textwidth}{!}{%
		\begin{tabular}{@{}lccccccccc cccccccc}
			\toprule
			\multicolumn{1}{l}{\multirow{2}{*}{Method}} & 
			\multicolumn{2}{c}{CIFAR B0 Inc5} & \multicolumn{2}{c}{CUB B0 Inc10} 
			& \multicolumn{2}{c}{IN-R B0 Inc20}
			& \multicolumn{2}{c}{IN-A B0 Inc20}
			& \multicolumn{2}{c}{ObjNet B0 Inc10}
			& \multicolumn{2}{c}{OmniBench B0 Inc30}
			& \multicolumn{2}{c}{VTAB B0 Inc10} \\
			& {$\bar{\mathcal{A}}$} & ${\mathcal{A}_B}$  
			& {$\bar{\mathcal{A}}$} & ${\mathcal{A}_B}$
			& {$\bar{\mathcal{A}}$} & ${\mathcal{A}_B}$   
			& {$\bar{\mathcal{A}}$} & ${\mathcal{A}_B}$ 
			& {$\bar{\mathcal{A}}$} & ${\mathcal{A}_B}$ 
			& {$\bar{\mathcal{A}}$} & ${\mathcal{A}_B}$ 
			& {$\bar{\mathcal{A}}$} & ${\mathcal{A}_B}$ 
			\\
			\midrule
			Finetune	& 38.90 & 20.17 &26.08 & 13.96 &32.31 &22.78  &24.28 & 14.51 & 19.14 & 8.73 & 23.61 & 10.57 & 34.95 & 21.25  \\
			Finetune Adapter & 60.51 &49.32& 66.84 &52.99 &58.17  &52.39 &45.41 &41.10 &50.22 &35.95 &62.32& 50.53 &48.91 & 45.12 \\
			LwF& 46.29 & 41.07 &48.97 & 32.03  &45.72  &34.17 &37.75 & 26.84 & 33.01 & 20.65 & 47.14 &33.95 & 40.48 & 27.54\\
			L2P   & 85.94 & 79.93 &67.05 & 56.25 & 75.46 & 69.77 &  49.39 & 41.71 &  63.78 & 52.19 &73.36 & 64.69 & 77.11 & 77.10\\
			DualPrompt    &87.87 & 81.15& 77.47 & 66.54 &73.10 &67.18  & 53.71 & 41.67 & 59.27 & 49.33 & 73.92 & 65.52 & 83.36 & 81.23\\
			CODA-Prompt & 89.11 & 81.96 & 84.00 & 73.37 & 77.97 &72.27 & 53.54 & 42.73 & 66.07 &53.29 &77.03 &68.09 &83.90 &83.02\\
			SimpleCIL   &  87.57 & 81.26 & 92.20 & 86.73 &61.26 &54.55  & 59.77 & 48.91 & 65.45 & 53.59 & 79.34 & 73.15 & 85.99 & 84.38\\
			\APER+ Finetune   & 87.67 & 81.27 & 91.82 & 86.39 &68.54 &58.37  &   61.01 & 49.57 & 61.41 & 48.34 & 73.02 & 65.03 &  87.47 & 80.44\\
			\APER+ VPT-S    &  90.43 & 84.57& 92.02 &86.51 &68.83  &62.03 & 58.39 & 47.20& 64.54 & 52.53 & 79.63 & 73.68 & 87.15 &  85.36\\
			\APER + VPT-D & 88.46 & 82.17 & 91.02 &84.99 &77.05  & 69.47 & 58.48 & 48.52 & 67.83 & 54.65 &  81.05 &  74.47 & 86.59 & 83.06\\
			\APER + SSF  & 87.78 & 81.98   & 91.72 &86.13&75.47  &67.02  &61.30 & 50.03 & 69.15 & 56.64 &  80.53 & 74.00 & 85.66 & 81.92\\
			\APER + Adapter &   90.65 &  85.15 &92.21 &86.73 &75.82  &67.95 & 60.47 &49.37 &  67.18 & 55.24 &  80.75 & 74.37 &  85.95 & 84.35\\
                SLCA &92.49  & 88.55   & 89.51 &82.19 &81.17  &77.00  &68.66  & 58.74& 72.55  & 61.30 & 82.80 &74.10  & 90.94  &90.76 \\ 
                EASE & 91.51 &85.80 &92.23 &86.81 & 81.74 & 76.17 & 65.34 & 55.04 & 70.84 & 57.86 & 81.11 & 74.85 & \bf93.61 & \bf93.55 \\
			\midrule
                \rowcolor{LightCyan}
			\name & \bf93.30  & \bf89.25  & \bf93.49 & \bf90.12  & \bf82.96 &\bf77.93  & \bf69.13 & \bf59.12 &\bf74.69	& \bf63.62  &\bf85.91 &\bf80.05 &92.62 &92.79 \\
			\bottomrule
		\end{tabular}
	}
\end{table*}

\subsection{Self-Refined Adapter Retrieval Mechanism} \label{sec:refine}
After obtaining these task-specific adapters, we utilize a prototype-based classifier~\cite{snell2017prototypical} for prediction. Specifically, after the training process of each incremental stage, we extract the class prototype of the $i$-th class using adapter $\mathcal{A}_b$:
\begin{equation} \label{eq:prototype}
    \boldsymbol{p}_{i,b}=\frac1N\sum\nolimits_{j=1}^{|\mathcal{D}^b|}\mathbb{I}(y_j=i)\phi(\mathbf{x}_j;\mathcal{A}_b) ,
\end{equation}
where $N$ is the instance number of class $i$. Eq.~\ref{eq:prototype} illustrates the constrution of classifier.  
During inference, we directly adopt the class prototype as the classifier weight, \ie, $\boldsymbol{w_i}=\boldsymbol{p_i}$, and utilize a cosine classifier for classification: 
\begin{equation} \label{eq:final_fx}
f(\x|\mathcal{A}_i)=(\frac {W}{\|W\|_2})^\top(\frac{\phi(\x;\mathcal{A}_i)}{\|\phi(\x;\mathcal{A}_i)\|_2}),
\end{equation}
where $\mathcal{A}_i$ denotes the selected adapter for the input $\x$.

Eq.~\ref{eq:key-value} illustrates how prompts are selected from the prompt pool. Subsequently, L2P integrates the selected prompt into the original PTM (\ie, $\phi(\x;P)$) to guide the model's response. However, this approach heavily relies on the retrieval mechanism of key-query pairs. Mistakenly retrieving the irrelevant prompts often leads to performance decay.
To address the retrieval-level issue, we design the testing stage surgery which uses self-refined adapter retrieval mechanism. It is an efficient and training-free method that enables the model to autonomously correct this problem, thereby improving adapter retrieval. This mechanism does not require any additional training overhead and is only used during the inference process, making the algorithm both simple and efficient. \looseness=-1

Since there is a gap between the PTM and downstream datasets, we first use an adapter to fine-tune the PTM on the first incremental task, denoting the model as $f(\x;\mathcal{A}_1)$. This process effectively bridges this gap and makes the model suitable as the initial selector. During inference, we utilize $f(\x;\mathcal{A}_1)$ to obtain the embedding of each testing example and perform the initial retrieval of task-specific adapters. Specifically, given an input $\x$, we first obtain the prediction result $f(\x|\mathcal{A}_1)$ of the model through Eq.~\ref{eq:final_fx}. Afterwards, we can easily infer its corresponding task ID $i$: 
\begin{equation} \label{eq:get_task_id_i}
    i = \lfloor \frac{\text{argmax}(f(\mathbf{x}|\A_1))}{|Y_b|} \rfloor
\end{equation}
where $Y_b$ is the number of classes for each task. Building on this result, we introduce an iterative self-refined process. As defined in Eq.~\ref{eq:final_fx}, this process primarily uses $f(\x;\A_i)$ to obtain prediction and identify the task ID $j$. \emph{Since each adapter is task-specific, we can determine whether to end the iteration by checking if $i=j$.} Specifically, through $f(\x|\mathcal{A}_i)$, we can infer its corresponding task ID $j$:
\begin{equation} \label{eq:get_task_id_j}
    j = \lfloor \frac{\text{argmax}(f(\mathbf{x}|\A_i))}{|Y_b|} \rfloor
\end{equation}
For example, in a scenario where each task comprises 10 classes, classes 0 through 9 are in the task 0, while classes 10 through 19 are in the task 1. Subsequently, if $i \neq j$, we replace $i$ with $j$ and repeat the process of Eq.~\ref{eq:get_task_id_j} until $i = j$, ensuring the self-consistency. 

\noindent\textbf{Effect of self-refined adapter retrieval mechanism.} 
Figure~\ref{figure:teaser} (right) illustrates the self-refined process. Firstly, $\phi(\x;\A_1)$ with the prototype-based classifier bridges the gap between upstream and downstream datasets, enhancing the model's ability to generalize to new classes. Hence, we use it as the initial selector to start the self-refined iteration. Moreover, this approach is training-free and does not incur any additional training costs, ensuring the algorithm's efficiency. Due to the self-refined adapter retrieval mechanism allowing the model to verify the correctness of its initial predictions, we can easily check model consistency, thereby alleviating the aforementioned mistaken retrieval problem. By using this mechanism, \name successfully corrects some images that were originally incorrectly predicted. The detailed visualization examples will be provided in experiments. 

\subsection{Multi-Stage Model Ensemble}
Inspired by the \emph{Complementary Learning System} of the human brain~\cite{mcclelland1995there,kumaran2016learning}, which suggests that the anterior cingulate circuit is responsible for rapid pattern recognition and unconscious memory, and the hippocampal circuit for deep processing and conscious memory. Therefore, we implement a two-stage model ensemble:
\begin{equation} \label{eq:ensemble}
    y^* = \argmax_y \ (\underbrace{f(\x|\A_1)}_{\text{Part 1}} + \underbrace{f(\x|\A_j)}_{\text {Part 2}} ).
\end{equation}
In Eq.~\ref{eq:ensemble}, Part 1, trained solely on the first incremental task, acts as a crucial bridge between the upstream and downstream datasets. It not only demonstrates strong generalization but also has the ability to quickly recognize and update information. In contrast, Part 2 employs progressively merged adapters and a self-refined adapter retrieval mechanism for deep processing and conscious memory.

\begin{algorithm}[t]
    \small
    \caption{\name for CIL}
    \label{alg1}
    {\bf Input}: Incremental datasets: $\{\D^{1}, \D^{2}, \cdots, \D^{B}\}$, Testing datasets: $\{\D_t^{1}, \D_t^{2}, \cdots, \D_t^{B}\}$, Task-specific Adapters: $\{\mathcal{A}_1, \cdots, \mathcal{A}_B\}$;
    \begin{algorithmic}[1]
        \For {task $b=1,2\cdots,B$}  {\Comment{\color{cyan}{Training stage}}}
        \State Get the incremental training set $\D^b$;
        \State Optimize task-specific $\mathcal{A}_b$ and $W$ via Eq.~\ref{eq:adapter-opt};
        \State Merge Adapter $\mathcal{A}_b$ via Eq.~\ref{eq:adapter-merge}; {\Comment{\color{cyan}{Adapter merging}}}
        \State Extract the prototypes via Eq.~\ref{eq:prototype}; 
        
        \For{each $\x \in \D_t^b$} {\Comment{\color{cyan}{Testing stage}}}
        \State Obtain the initial prediction task ID $i$ via Eq.~\ref{eq:get_task_id_i}; \label{line:16}
        \State Correct adapter iteratively via Eq.~\ref{eq:get_task_id_j} {\Comment{\color{cyan}{Self-refined}}}
        \State Calculate the $y^*$ via Eq.~\ref{eq:ensemble}; \label{line:25}
        \EndFor
        \EndFor
    \end{algorithmic}
\end{algorithm}

\begin{figure*}[t]
	\centering
	\begin{subfigure}{0.3\linewidth}
		\includegraphics[width=1\columnwidth]{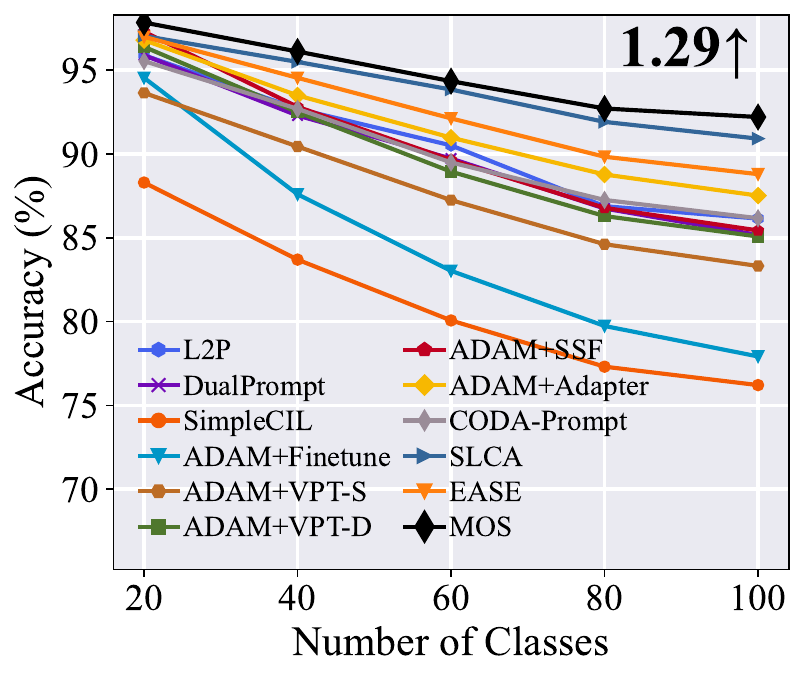}
		\caption{\small CIFAR B0 Inc20}
		\label{fig:benchmark-cifar}
	\end{subfigure}
	\hfill
	\begin{subfigure}{0.3\linewidth}
		\includegraphics[width=1\linewidth]{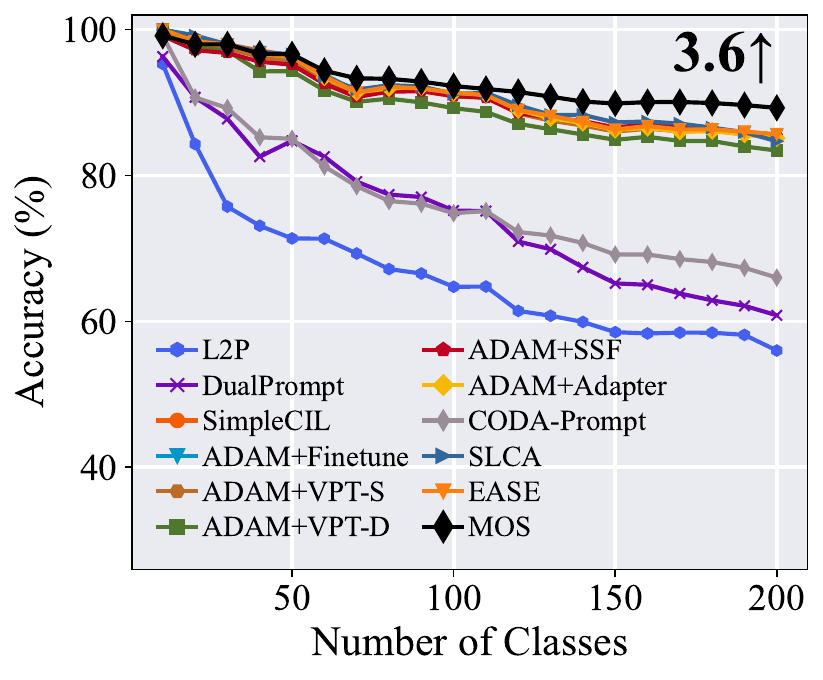}
		\caption{\small CUB B0 Inc10}
		\label{fig:benchmark-cub}
	\end{subfigure}
	\hfill
	\begin{subfigure}{0.3\linewidth}
		\includegraphics[width=1\linewidth]{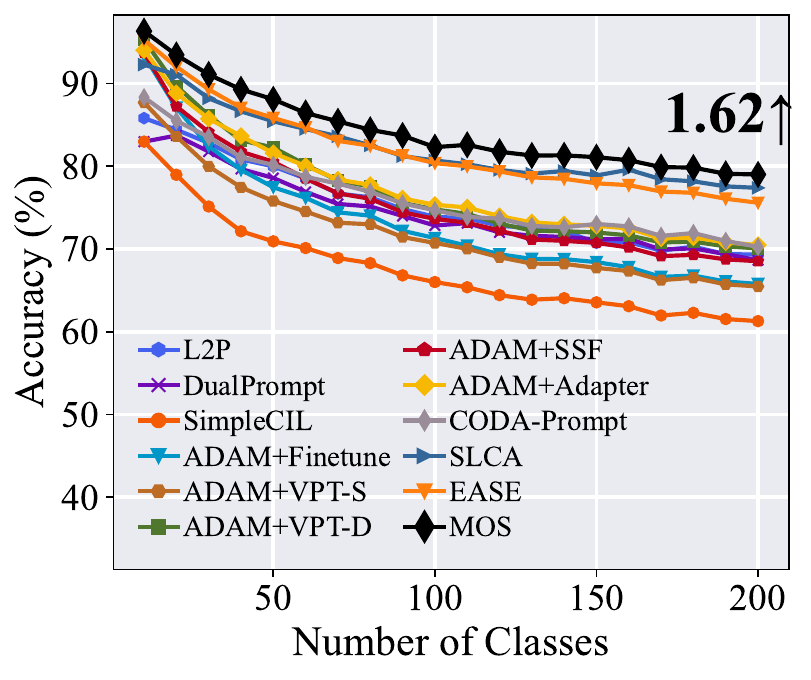}
		\caption{\small ImageNet-R B0 Inc10}
		\label{fig:benchmark-inr}
	\end{subfigure}
	\\
	\begin{subfigure}{0.3\linewidth}
		\includegraphics[width=1\linewidth]{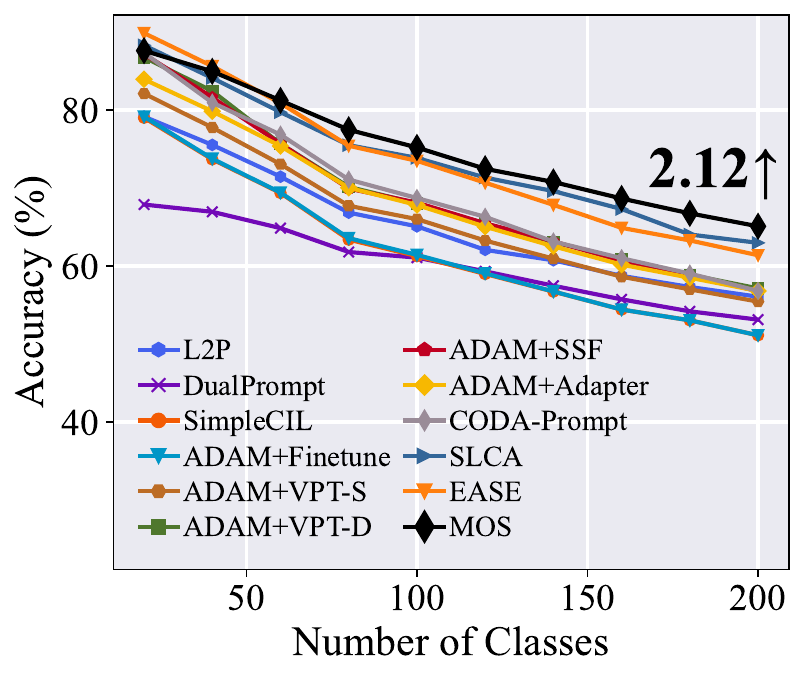}
		\caption{\small ObjectNet B0 Inc20}
		\label{fig:benchmark-obj}
	\end{subfigure}
	\hfill
	\begin{subfigure}{0.3\linewidth}
		\includegraphics[width=1\linewidth]{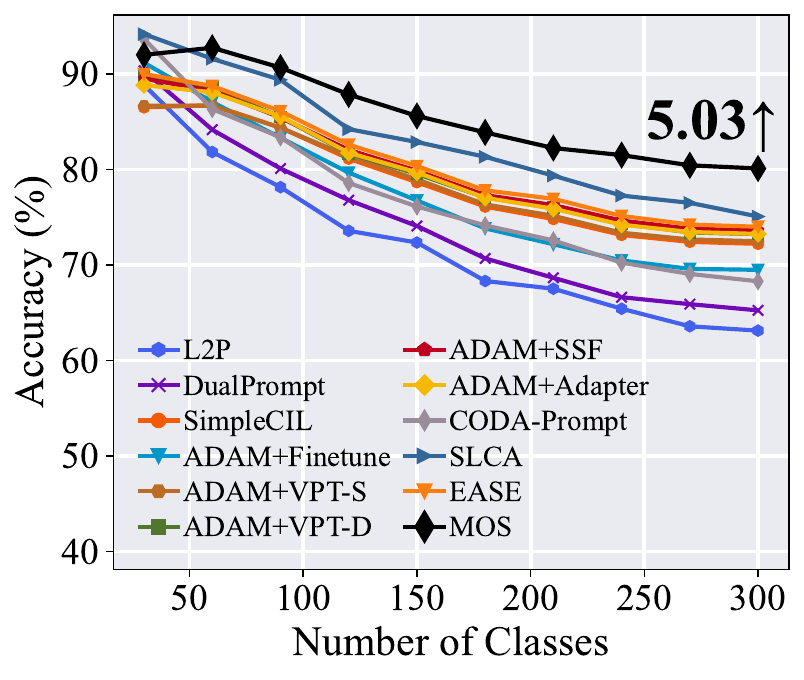}
		\caption{\small Omnibenchmark B0 Inc30}
		\label{fig:benchmark-omni}
	\end{subfigure}
	\hfill
	\begin{subfigure}{0.3\linewidth}
		\includegraphics[width=1\columnwidth]{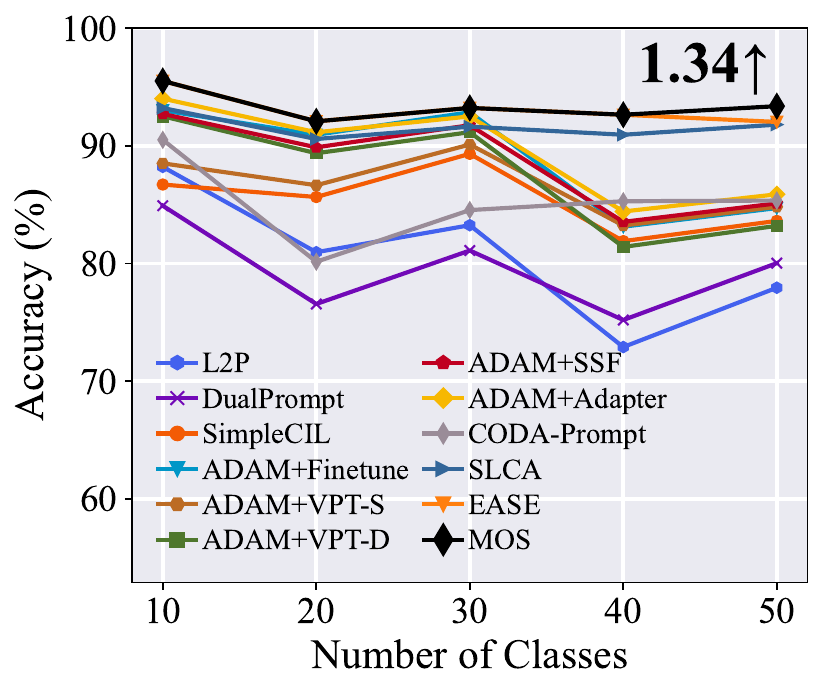}
		\caption{\small VTAB B0 Inc10}
		\label{fig:benchmark-vtab}
	\end{subfigure}
	\caption{\small Performance curve of different methods under different settings. All methods are initialized with {\bf ViT-B/16-IN1K}. We annotate the relative improvement of \name above the runner-up method with numerical numbers at the last incremental stage. }
	\label{fig:benchmark}
\end{figure*}

\noindent\textbf{Summary of \mame.} We summarize the detailed pipeline in Algorithm~\ref{alg1}. We initialize and train an adapter for each incremental task to encode the task-specific information, and employ the adapter merging strategy to mitigate parameter drift phenomenon. Subsequently, we extract prototypes from the current dataset for the current adapter to complete the classifier head. During inference, we utilize the training-free self-refined adapter retrieval mechanism to correct the mistaken retrieval of irrelevant adapters. Finally, we implement a two-stage model ensemble to select the maximum logit.\looseness=-1
\section{Experiments}
\label{sec:experiments}
In this section, we evaluate \name on seven benchmark datasets and compare it to other SOTA methods to demonstrate its superiority. Moreover, we provide an ablation study and a visualized analysis to validate the robustness of \mame.

\subsection{Implementation Details}
\noindent\textbf{Dataset:} Since PTMs possess extensive knowledge regarding upstream tasks, we follow \cite{zhou2024revisiting} to evaluate the performance on CIFAR100~\cite{krizhevsky2009learning}, CUB200~\cite{WahCUB2002011}, ImageNet-R~\cite{hendrycks2021many}, ImageNet-A~\cite{hendrycks2021natural}, objectNet~\cite{barbu2019objectnet}, Omnibenchmark~\cite{zhang2022benchmarking}, and VTAB~\cite{zhai2019large}. These datasets represent typical CIL benchmarks and include out-of-distribution datasets that exhibit a \emph{significant domain gap} with ImageNet (\ie, the pre-trained dataset). Specifically, There are 50 classes in VTAB, 100 classes in CIFAR100, 200 classes in CUB, ImageNet-R, ImageNet-A, ObjectNet, and 300 classes in OmniBenchmark. More details are reported in the supplementary. \looseness=-1

\noindent\textbf{Dataset split:} Following the benchmark setting~\cite{rebuffi2017icarl,wang2022learning}, we utilize the notation `B-$m$ Inc-$n$' to represent class splits, where $m$ indicates the number of classes in the initial task, and $n$ denotes the number of classes in each subsequent incremental task. $m=0$ means the total classes are equally divided into each task. For a consistent and fair comparison, we randomly shuffle class orders using a random seed of 1993 before splitting the data. 
We ensure consistency in the training and testing sets across all methods, following ~\cite{zhou2024revisiting}.

\noindent\textbf{Training details:} We use PyTorch~\cite{paszke2019pytorch} and PILOT~\cite{sun2023pilot} to implement all models on NVIDIA RTX 4090 with the \emph{same} network backbone. Since the wide range of PTMs are publicly accessible~\cite{rw2019timm}, we choose two representative models following~\cite{wang2022dualprompt,zhou2024revisiting}, denoted as \textbf{ViT-B/16-IN1K} and \textbf{ViT-B/16-IN21K}. They are both initially pre-trained on ImageNet21K, while the former is further finetuned on ImageNet1K. In \mame, we set the batch size to 48 and train for 20 epochs using the SGD optimizer with momentum. The learning rate is initially set to 0.01 and follows a cosine annealing decay pattern. The projection dimension $r$ in the adapter is set to 16, and the EMA factor parameter $\alpha$ is set to 0.1.

\noindent\textbf{Comparison methods:} We choose state-of-the-art PTM-based CIL methods for comparison, such as Finetune Adapter~\cite{chenadaptformer}, L2P~\cite{wang2022learning}, DualPrompt~\cite{wang2022dualprompt}, CODA-Prompt~\cite{smith2023coda}, SimpleCIL~\cite{zhou2024revisiting}, APER~\cite{zhou2024revisiting}, SLCA~\cite{zhang2023slca}, EASE~\cite{zhou2024expandable}. In addition, we compare \name with traditional CIL methods modified by PTM, including LwF~\cite{li2017learning}, FOSTER~\cite{wang2022foster}, MEMO~\cite{zhou2022model}, iCaRL~\cite{rebuffi2017icarl}, DER~\cite{yan2021dynamically}. We report the baseline method, which sequentially finetunes the PTM, denoted as Finetune. All methods are implemented with the \textbf{same} PTM for a \emph{fair} comparison.

\noindent\textbf{Evaluation protocol:} Following the benchmark established by~\cite{rebuffi2017icarl}, we denote the Top-1 accuracy after the $b$-th stage as $\mathcal{A}_b$. Moreover, we use $\mathcal{A}_B$ (the performance after the last stage) and $\bar{\mathcal{A}}=\frac{1}{B}\sum_{b=1}^{B}\mathcal{A}_b$ (average performance along incremental stages) as measurements.

\subsection{Benchmark Comparison}
In this section, we compare \name with other SOTA methods across seven datasets and various backbone weights. As detailed in Table~\ref{tab:benchmark}, \name shows the best performance across all seven benchmarks, significantly surpassing the SOTA methods, such as SLCA, EASE, and APER. Furthermore, we present an analysis of the incremental performance trend of different methods in Figure~\ref{fig:benchmark} with ViT-B/16-IN1K. Notably, \name outperforms the runner-up method by 2\%$\sim$5\% on CUB, ObjectNet, and OmniBenchmark, as highlighted in the annotations at the end of each image.

\begin{figure}[t]
	\centering
	\begin{subfigure}{0.49\linewidth}
		\includegraphics[width=1\columnwidth]{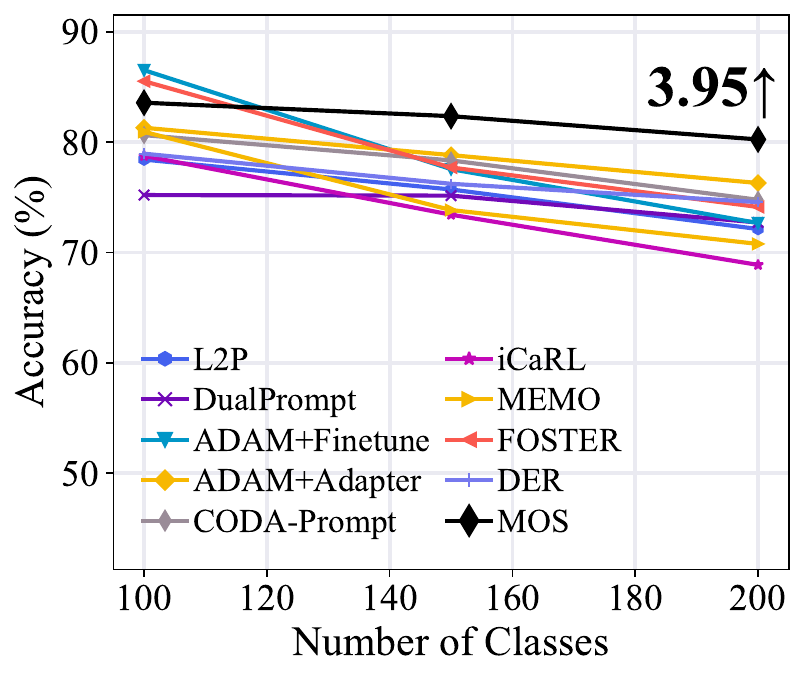}
		\caption{\small ImageNet-R B100 Inc50}
		\label{fig:benchmark-inrb100inc50}
	\end{subfigure}
	\hfill
	\begin{subfigure}{0.49\linewidth}
		\includegraphics[width=1\linewidth]{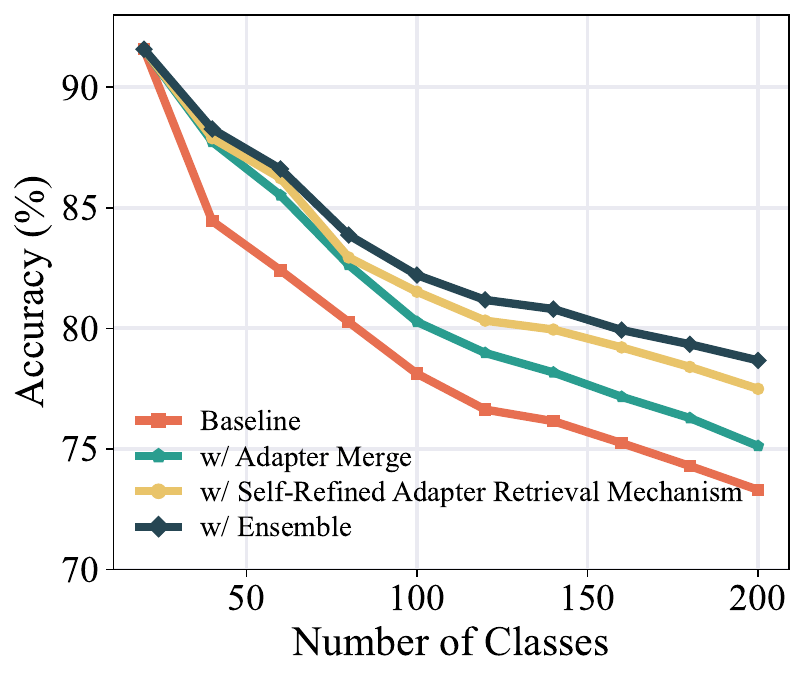}
		\caption{\small Ablation study}
		\label{figure:ablation}
	\end{subfigure}
	\caption{\small \textbf{Left:} Experimental results with large base classes. All methods are based on the same PTM ({\bf ViT-B/16-IN1K}). \textbf{Right:} Ablation study of different components in \mame. We find each component within \name enhances the performance.}
	\label{fig:benchmark-large-base}
\end{figure}

Beyond the B0 setting presented in Table~\ref{tab:benchmark} and Figure~\ref{fig:benchmark}, we extend our experiments to a larger base setting. In Figure~\ref{fig:benchmark-inrb100inc50}, we compare \name with several SOTA methods and traditional methods using the \textbf{same} PTM. Although traditional methods require storing exemplars to recover previous knowledge, \name achieves SOTA performance in this setting as well. Extensive experiments validate the effectiveness of \mame. \looseness=-1

\begin{figure}[t] 
	\begin{center}
		{	
			\includegraphics[width=.95\columnwidth]{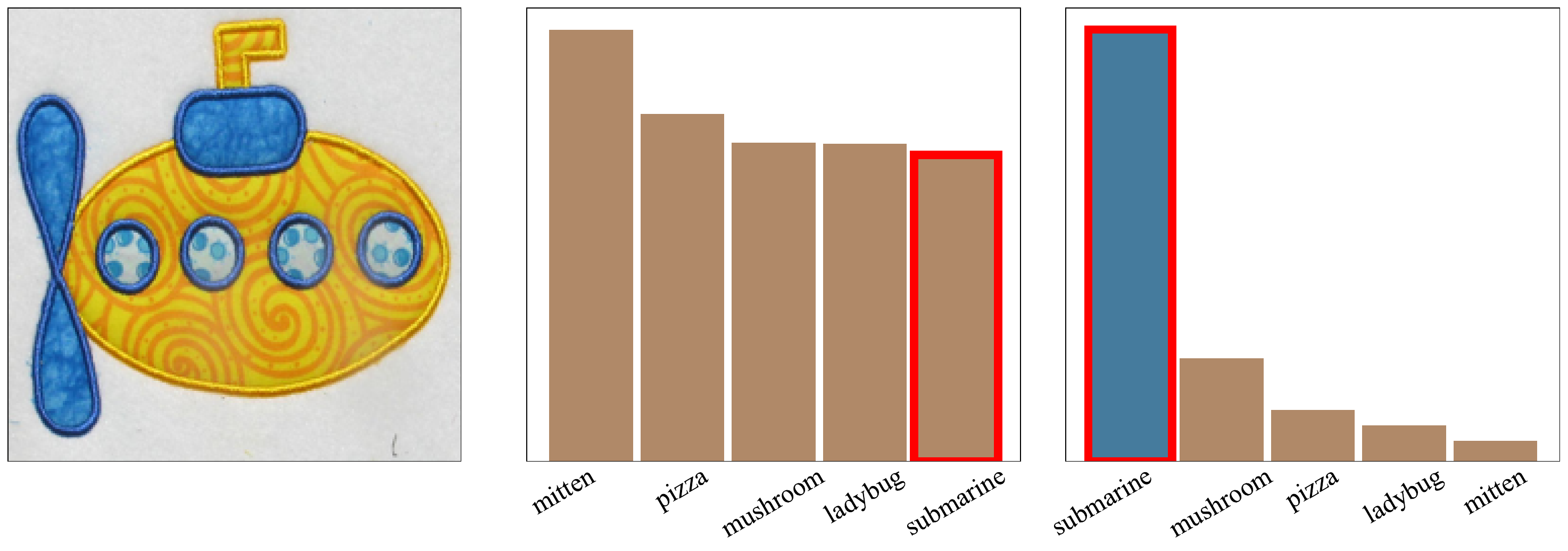}
			\includegraphics[width=.95\columnwidth]{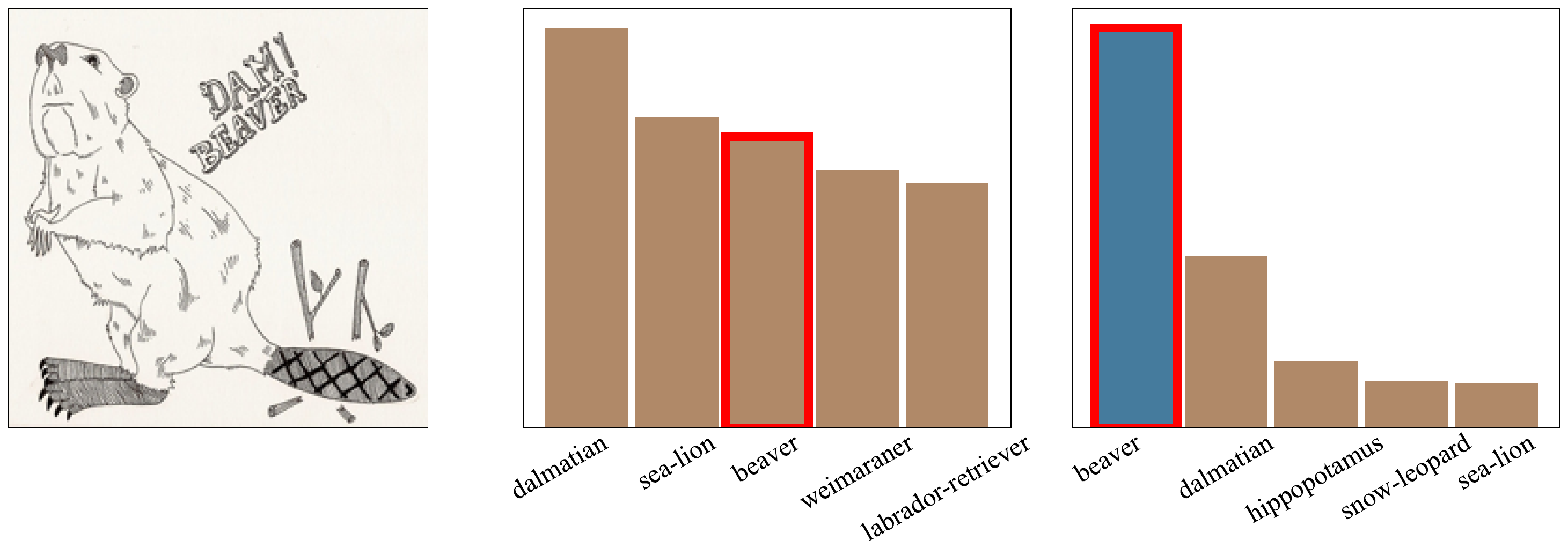}
			\includegraphics[width=.95\columnwidth]{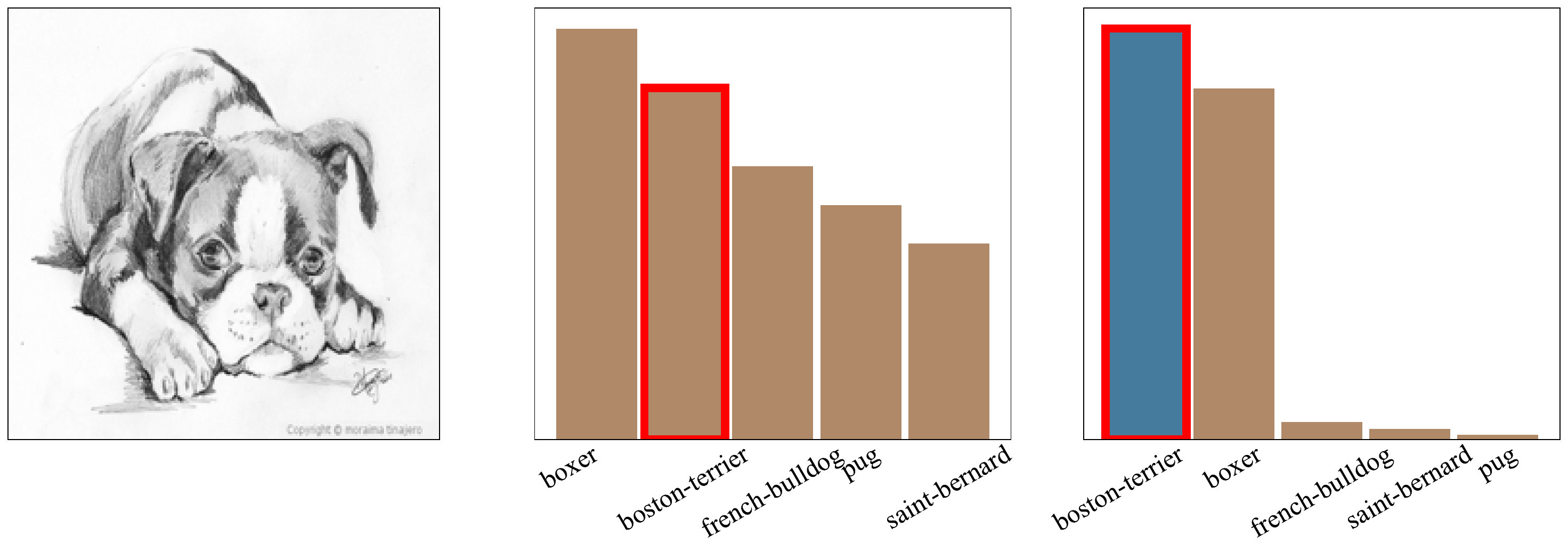}
			\includegraphics[width=.95\columnwidth]{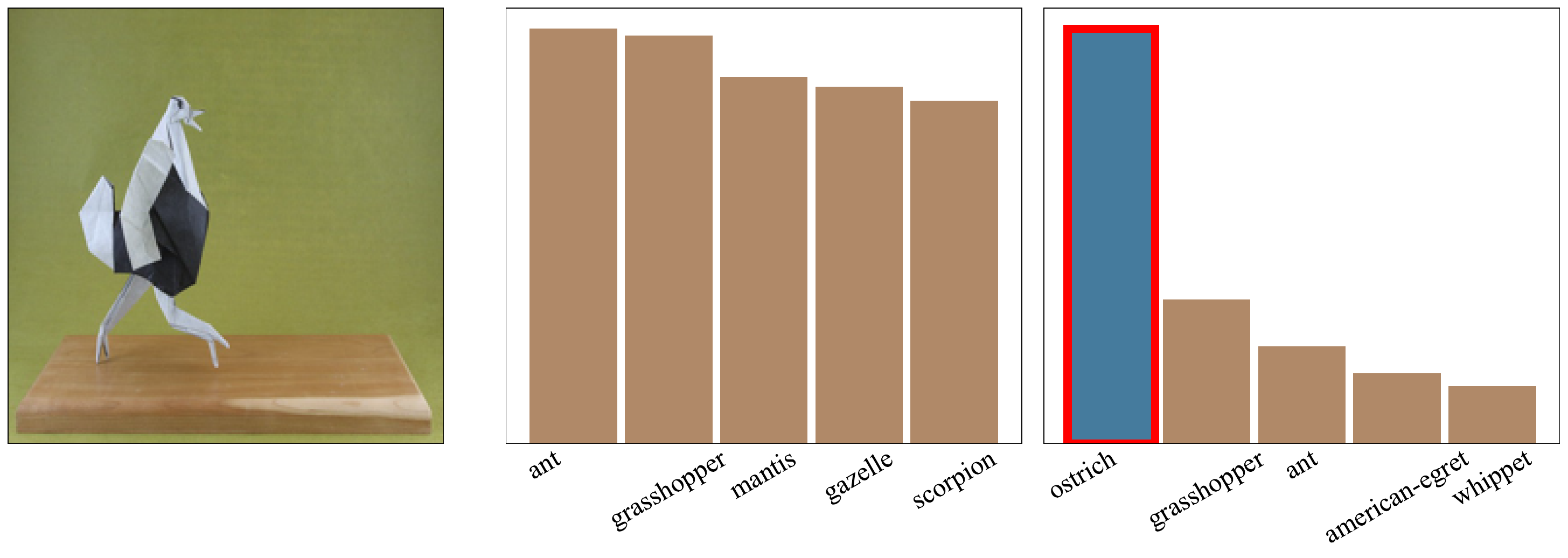}
		}
		
	\end{center}
	\caption{ \small Visualizations of self-refined adapter retrieval mechanism on ImageNet-R. The original images are depicted in the first row, followed by the top-5 prediction probability before the self-refined process, and the probabilities after refinement in the last row. The ground-truth class is highlighted with red boxes.
	} \label{figure:visual_self-refine}
\end{figure}

\subsection{Ablation Study}
In this section, we conduct an ablation study by \textit{incrementally adding each component} to evaluate their effectiveness within \mame. Specifically, we present this ablation study on ImageNet-R B0 Inc20 setting. As depicted in Figure~\ref{figure:ablation}, \textbf{`Baseline'} refers to the PTM integrated with $\mathcal{A}_1$ (\ie, $\phi(\x|\A_1)$). 
Since we aim to mitigate parameter drift and build task-specific adapters, we report \textbf{`w/ Adapter Merge'} by only using Eq.~\ref{eq:adapter-merge}. 
Due to the aforementioned mistaken retrieval issue, we propose using the model's inherent capabilities to correct errors. We report the performance of \textbf{`w/ Self-Refined Adapter Retrieval Mechanism'} by using this technique along with the adapter merging strategy.
As shown in the figure, both the adapter merging strategy and self-refined adapter retrieval mechanism significantly improve the performance, which indicates \name has the ability to correct itself and alleviate the catastrophic forgetting. 
Finally, we adjust the logits using Eq.~\ref{eq:ensemble} to trade off stability and plasticity, denoted as \textbf{`w/ Ensemble'}. Ablations verify that every component in \name contributes to improving performance.

\subsection{Visualizations} 
\label{sec:visualize}

In this section, we discuss how the self-refined adapter retrieval mechanism works. To illustrate this, we present the visualization of prediction results before and after the self-refined process and analyze their differences. We choose images from ImageNet-R and utilize the model trained under the B0 Inc20 setting. The results are shown in Figure~\ref{figure:visual_self-refine}. As shown in these figures, \name is capable of rectifying incorrect predictions. This is evident even in the example below, where the initial top-5 class predictions do not include the ground truth, yet \name accurately corrects this error. It demonstrates that the model, using its inherent capabilities, can select the most suitable adapter for the current sample. Hence, \name can use this adapter to extract more suitable features, which aids in enhancing prediction accuracy. These visualizations reveal that the self-refined adapter retrieval mechanism can help to correct the outputs, thereby enhancing the attention of the ground-truth class. 
\section{Conclusion}
Incremental learning is an increasingly prominent paradigm in real-world systems. 
This paper proposes a novel model surgery (\mame) for PTM-based CIL to rescue the model from forgetting previous knowledge. Specifically, we introduce an adapter merging method to mitigate parameter drift and design a training-free self-refined adapter retrieval mechanism for better adapter retrieval during inference. Our approach balances the stability-plasticity dilemma by leveraging the model's inherent capabilities, enhancing generalization and adaptability. Extensive experiments on seven benchmark datasets validate the effectiveness of \mame. In future work, we aim to explore further application scenarios, such as few-shot class-incremental learning.

\section{Acknowledgments}
This work is partially supported by Fundamental Research Funds for the Central Universities (2024300373, 14380021), Key Program of Jiangsu Science Foundation (BK20243012), CCF-Tencent Rhino-Bird Open Research Fund RAGR20240101,
NSFC (62476123, 62376118, 62006112, 62250069, 61921006,  62402430), the AI \& AI for Science Project of
Nanjing University, Collaborative Innovation Center of Novel Software Technology and Industrialization.

\bibliography{aaai25}

\section{Appendix}
In this supplementary section, we present additional information on \mame, encompassing more details on implementation and expanded experimental results. The supplementary material is organized as follows:
\begin{itemize}

    \item Section 1\ref{sec:1} introduces further analysis of \mame, including parameter sensitivity, multiple runs, running time comparison, and more visualizations of the self-refined adapter retrieval mechanism.
    \item Section 2\ref{sec:2} presents more details of classifier alignment using Gaussian distribution.
    \item Section 3\ref{sec:3} offers more introduction of compared methods.
    \item Section 4\ref{sec:4} provides supplementary results of benchmark datasets to the main paper.       
\end{itemize}

\section{Further Analysis} \label{sec:1}
In this section, we conduct further analysis on \mame's components to investigate their effectiveness, \eg, parameter sensitivity, multiple runs, and more visualizations of self-refined mechanism. Additionally, we also compare \name with other methods on running time.

\subsection{Parameter Sensitivity}
\begin{figure}[t]
	\begin{center}
		{\includegraphics[width=1\columnwidth]{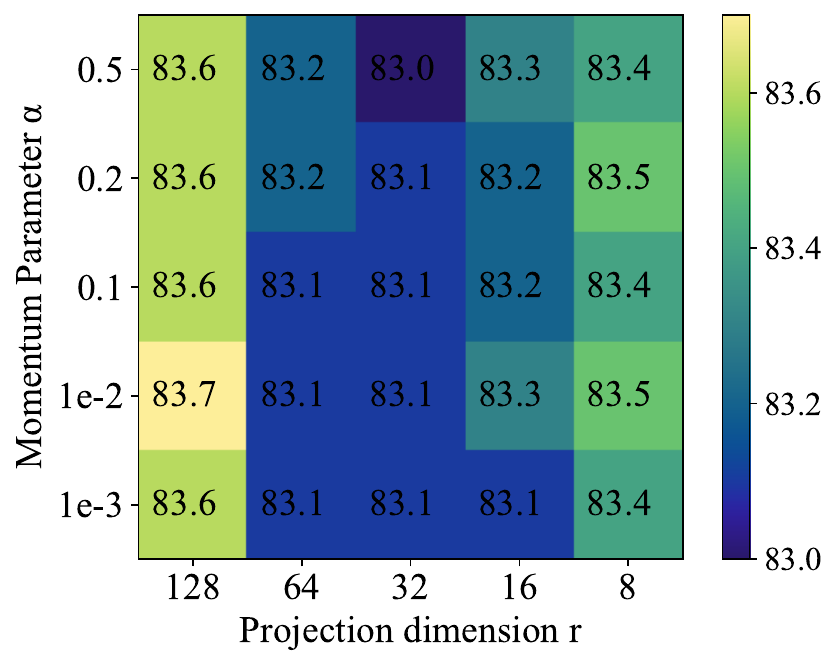}}
	\end{center}
	\caption{ Sensitivity of hyperparameters.}
	\label{fig:hyper_mean}
\end{figure}

In the main paper, we introduce progressively merged adapters with two key hyperparameters: the projection $r$ within the adapter and the merging momentum $\alpha$ utilized in the Exponential Moving Average (EMA) method. To evaluate the sensitivity of these parameters, we conducted experiments on the ImageNet-R B0 Inc20 dataset. Specifically, we varied $r$ over the set $\{8, 16, 32, 64, 128\}$ and $\alpha$ over $\{0.001, 0.01, 0.1, 0.2, 0.5\}$. The average performance across these settings is depicted in Figure~\ref{fig:hyper_mean}. As illustrated, the model's performance remains stable across a range of parameter values. Moreover, we can infer that the parameters are not highly sensitive. Based on these findings and with the possibility of reducing the number of parameters, we recommend default settings of $r=16, \alpha=0.1$ for other datasets.

\subsection{Multiple Runs}
\begin{figure}[t]
	\begin{center}
		{\includegraphics[width=0.9\columnwidth]{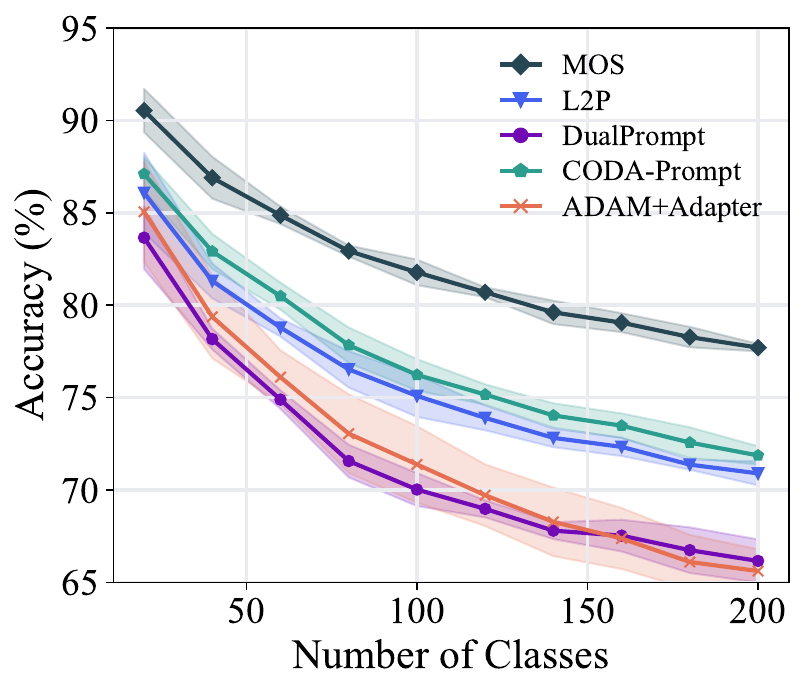}}
	\end{center}
	\caption{\small Results on ImageNet-R B0 Inc20 with multiple runs. \name \textbf{consistently outperforms other methods by a significant margin}.
	}
	\label{fig:multi-seeds}
\end{figure}

In the main paper, we conduct experiments across various datasets, following~\cite{rebuffi2017icarl} to randomize class orders using the seed $1993$. This section extends that work by repeating these experiments with multiple random seeds, specifically $\{1993,1994,1995,1996,1997\}$. This approach yields five sets of incremental results for different methods, allowing us to calculate and present the mean and standard deviation in Figure~\ref{fig:multi-seeds}. 

As depicted in the figure, we can infer that \name consistently surpasses other methods by a significant margin across a range of random seeds.

\subsection{Running Time Comparison}
\begin{figure}[t]
	\begin{center}
		{\includegraphics[width=0.9\columnwidth]{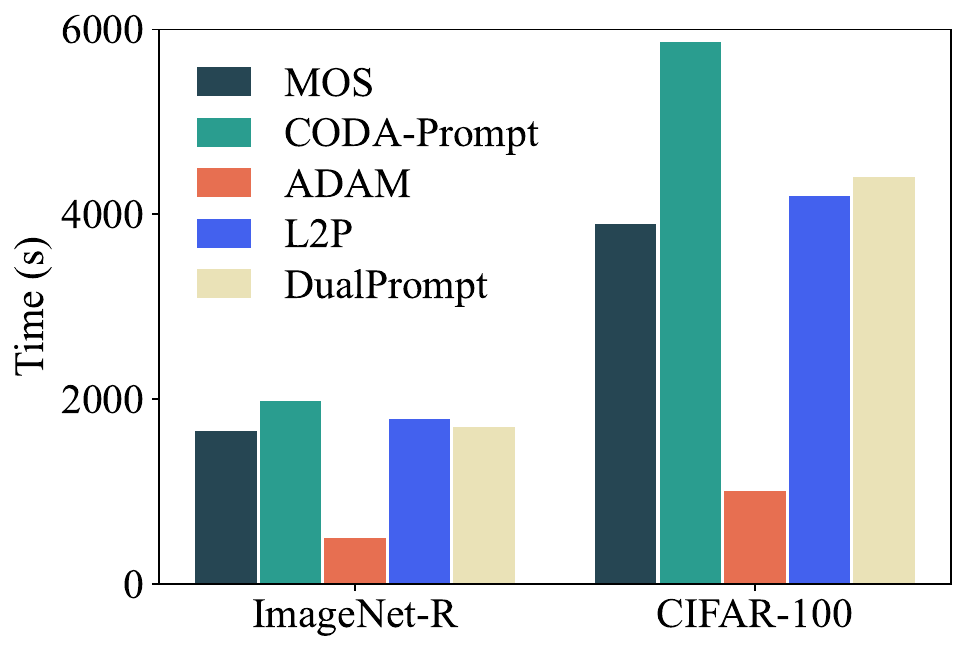}}
	\end{center}
	\caption{\small Running time comparison of different methods. \name \textbf{utilizes less running time than CODA-Prompt, L2P, and DualPrompt while having better performance}.
	}
	\label{fig:run-time}
\end{figure}

In this section, we present the comparative running time of various class-incremental learning methods. All experiments are conducted on a single NVIDIA 4090 GPU. Specifically, we train all methods for 10 epochs in ImageNet-R and 20 epochs in CIFAR-100. The outcomes are shown in Figure~\ref{fig:run-time}. The results indicate that \name outperforms CODA-Prompt, L2P, and DualPrompt in terms of running time while concurrently achieving superior performance. These results verify the efficacy of \name.

\begin{figure}[t]
    \centering
    \begin{subfigure}{0.85\linewidth}
        \includegraphics[width=1\columnwidth]{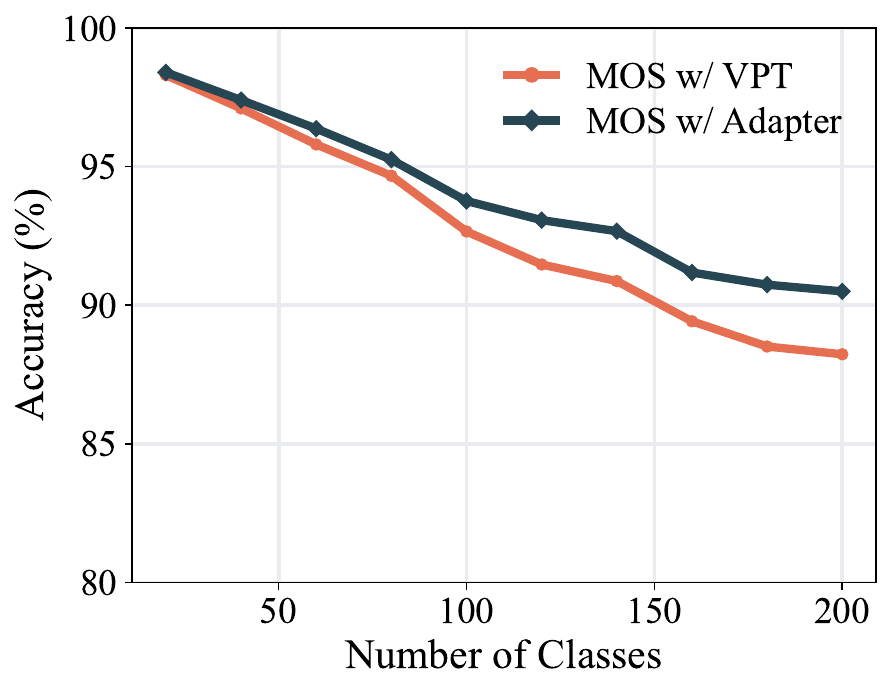}
        \caption{CIFAR100 B0 Inc10}
    \end{subfigure}
    \vfill
    \begin{subfigure}{0.85\linewidth}
        \includegraphics[width=1\columnwidth]{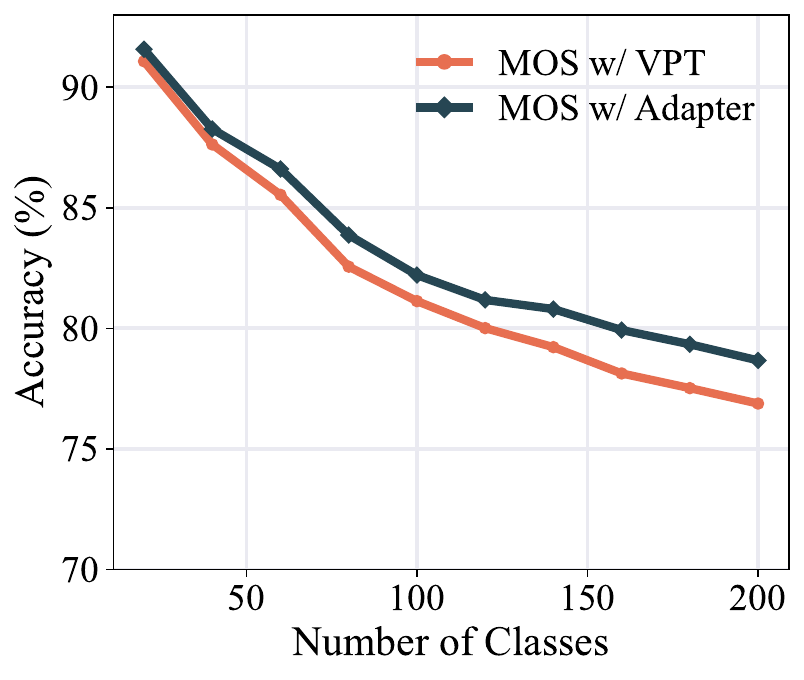}
        \caption{ImageNet-R B0 Inc20}
    \end{subfigure}
    \caption{ Experimental results on different subspace tuning methods. \textbf{Using adapter tuning shows better performance than VPT}.}
    \label{adapter-vs-vpt}
 \end{figure}

\subsection{More Visualizations}
\begin{figure*}[t] 
    \centering
    \begin{subfigure}{0.49\linewidth}
        \includegraphics[width=1\columnwidth]{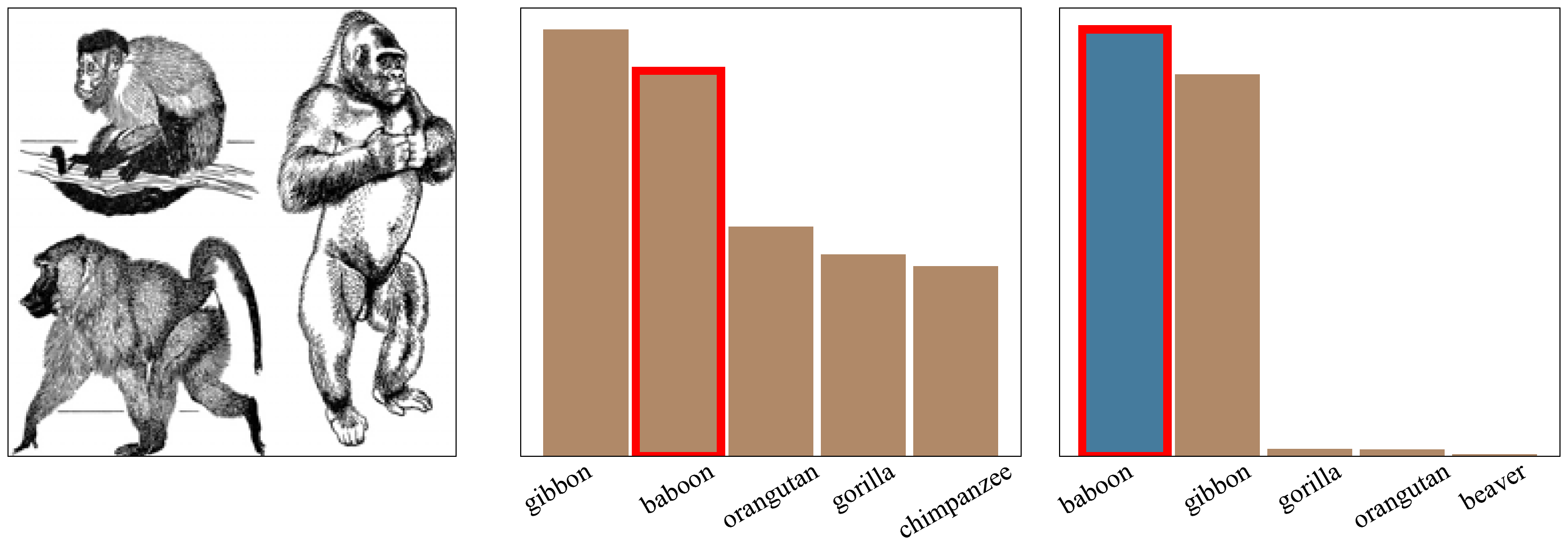}
    \end{subfigure}
    \hfill
    \begin{subfigure}{0.49\linewidth}
        \includegraphics[width=1\columnwidth]{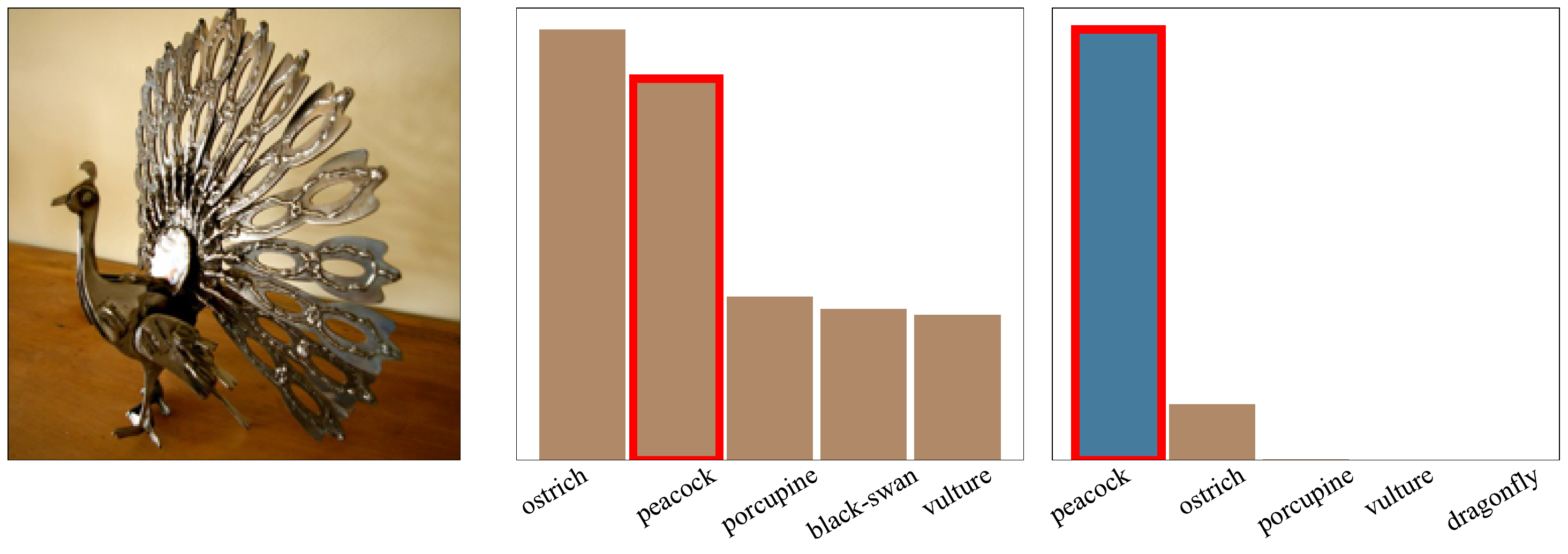}
    \end{subfigure}
    \\
    \begin{subfigure}{0.49\linewidth}
        \includegraphics[width=1\columnwidth]{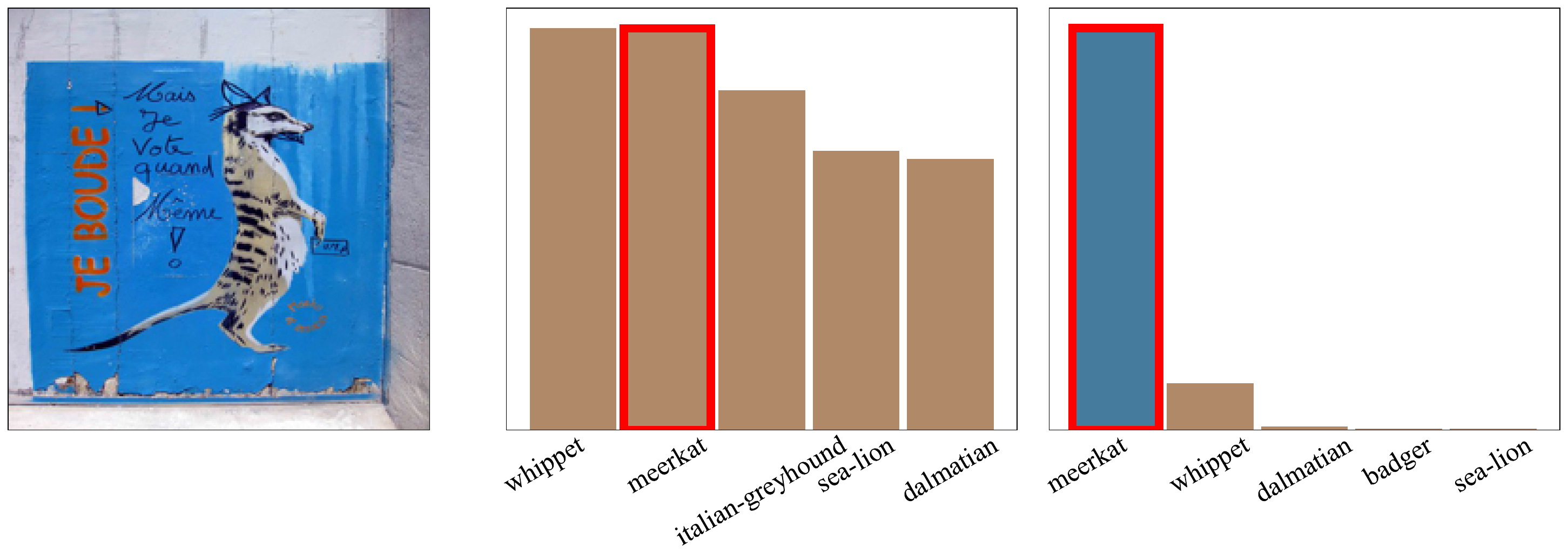}
    \end{subfigure}
    \hfill
    \begin{subfigure}{0.49\linewidth}
        \includegraphics[width=1\columnwidth]{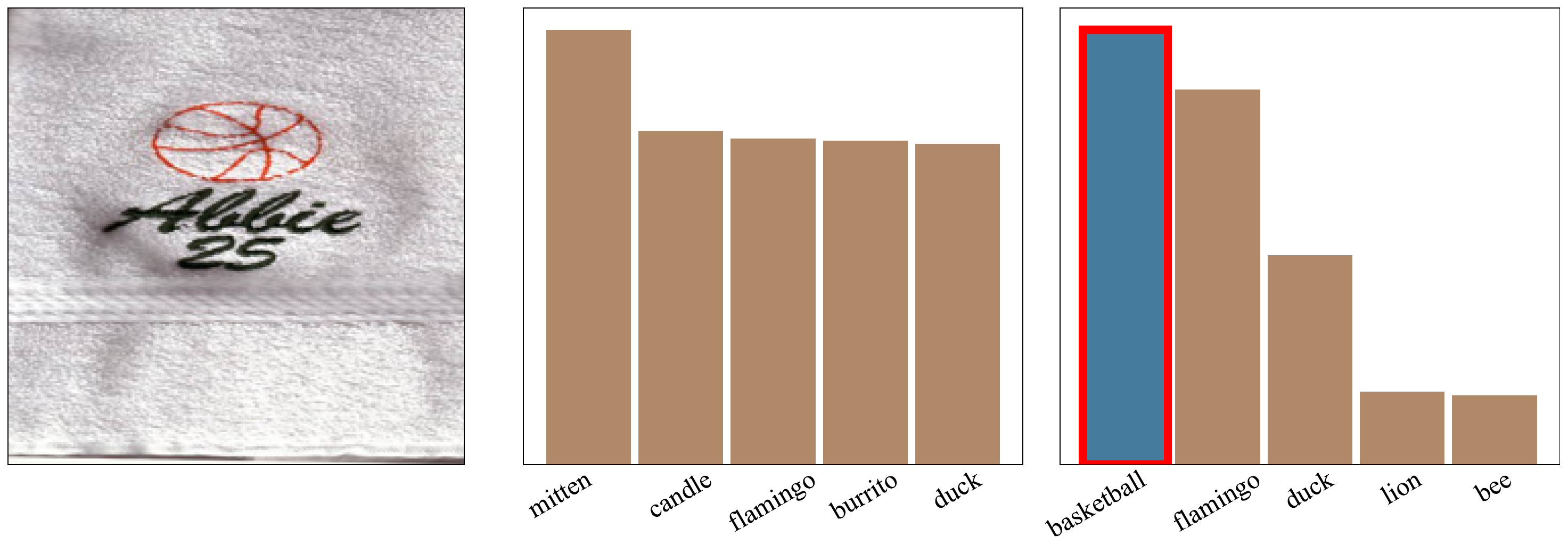}
    \end{subfigure}
    \\
    \begin{subfigure}{0.49\linewidth}
        \includegraphics[width=1\columnwidth]{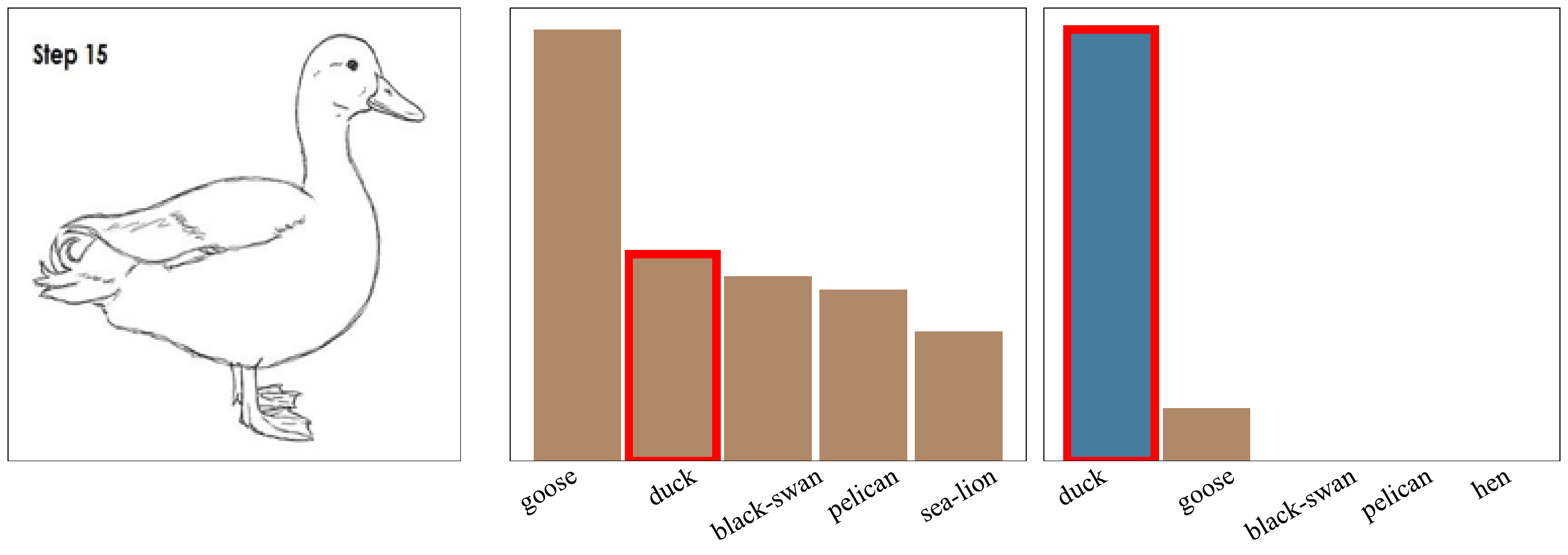}
    \end{subfigure}
    \hfill
    \begin{subfigure}{0.49\linewidth}
        \includegraphics[width=1\columnwidth]{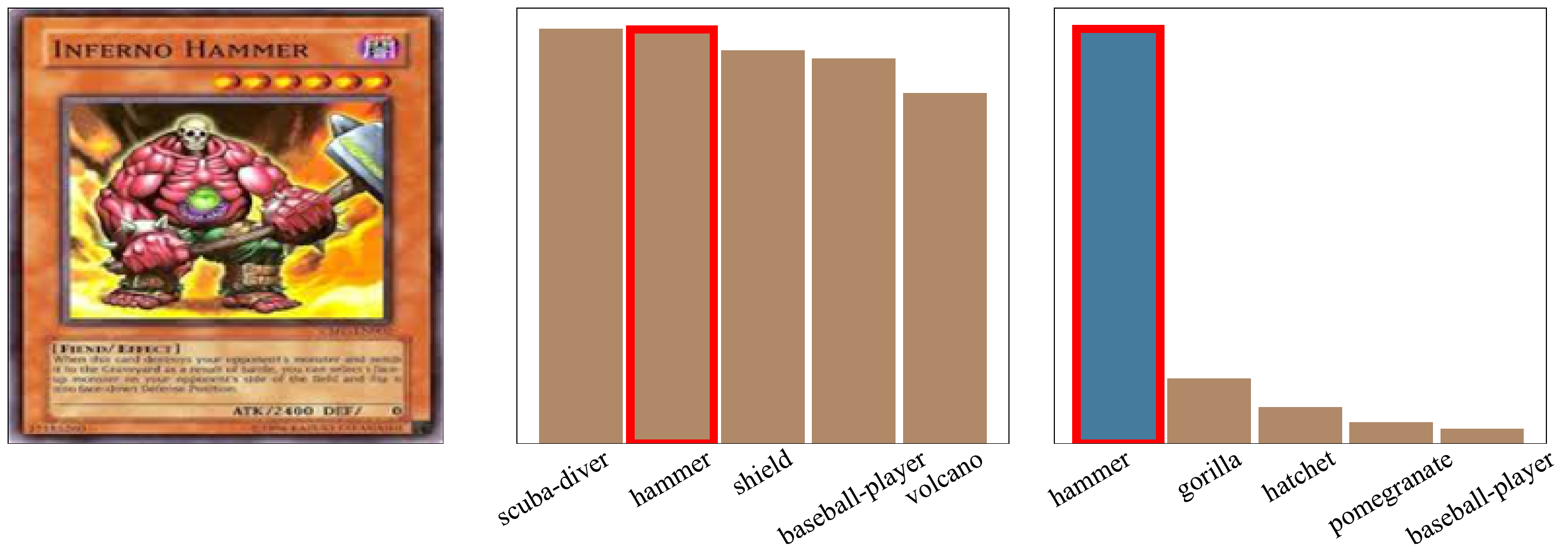}
    \end{subfigure}
    \\
    \begin{subfigure}{0.49\linewidth}
        \includegraphics[width=1\columnwidth]{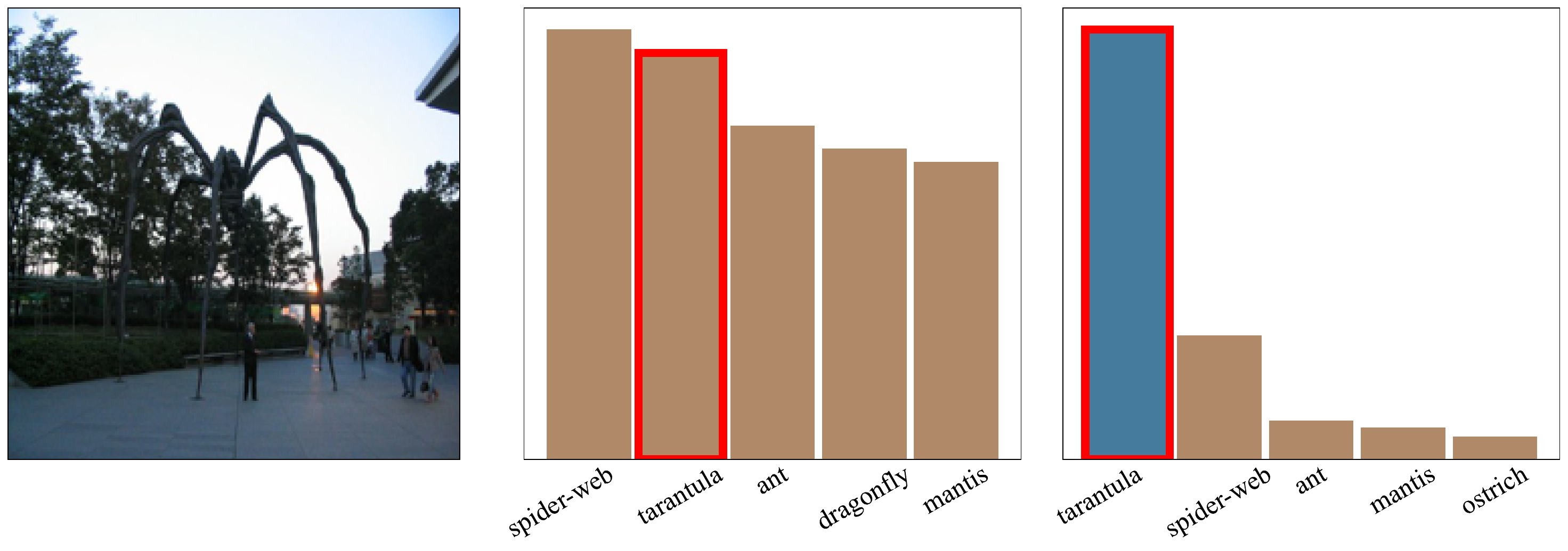}
    \end{subfigure}
    \hfill
    \begin{subfigure}{0.49\linewidth}
        \includegraphics[width=1\columnwidth]{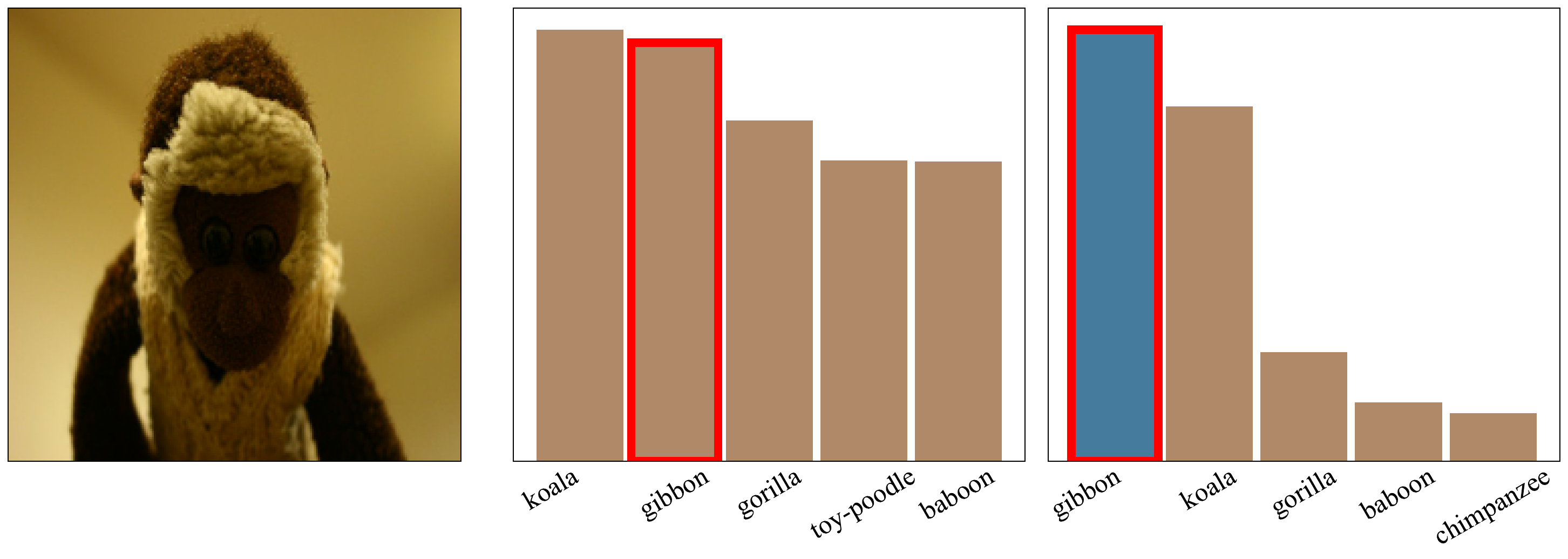}
    \end{subfigure}
		
	\caption{ More Visualizations of self-refined mechanism on ImageNet-R. The original images are depicted in the first row, followed by the top-5 prediction probability before the self-refined process in the second row, and the probabilities post-refinement in the last row. The ground-truth class is shown with red edges.
	} \label{fig:more-visual}
\end{figure*}

In the main paper, the functioning of the self-refined mechanism is elucidated through four visualizations. To further intuitively demonstrate the effectiveness of this method, additional visualizations are provided. Specifically, we choose images from ImageNet-R and utilize the model trained under the B0 Inc20 setting with ViT-B/16-IN1K. Further results are depicted in Figure~\ref{fig:more-visual}. These figures illustrate \mame's ability to rectify incorrect predictions. Moreover, the visualizations highlight how the self-refined mechanism aids in correcting outputs and enhances focus on the ground-truth class.

\section{Adapter Tuning VS. VPT}
In the main paper, we build task-specific components via adapter tuning~\cite{rebuffi2017learning}. However, besides adapter tuning, there are other methods to efficiently tune pre-trained models with parameters, such as visual prompt tuning~\cite{jia2022visual} (VPT). In this section, we integrate our method with various parameter-efficient fine-tuning (PEFT) techniques and conduct experiments on CIFAR100 and ImageNet-R. We maintain consistent setting and solely vary the PEFT training methods, and present the results in Figure~\ref{adapter-vs-vpt}.

From this figure, we observe that using adapters for model surgery achieves better performance than using VPT, surpassing it by $\sim$2$\%$ on these datasets. This superiority stems from two main aspects: firstly, adapter tuning exhibits stronger tuning capability than VPT. Secondly, adapter tuning only requires learning a set of adapters for each task, whereas VPT needs constructing a large prompt pool, complicating retrieval. Therefore, we choose the adapter tuning as the implementation of model surgery in \mame.

\section{Details of Examples Generation} \label{sec:2}
In this section, we provide a detailed explanation of how Gaussian distribution is utilized for aligning the classifier. Since the representations of PTM are typically well-distributed, after training each task-specific $\mathcal{A}_i$, we extract the mean ($\mu_c$) and variance ($\Sigma_c$) of features for each training category. These are then recovered by adding Gaussian noise to the Gaussian distribution. It enables the model to mitigate the bias~\cite{zhao2020maintaining} introduced to the classifier after learning the task-specific adapter at each stage, thereby facilitating the alignment of the classifier. 

First of all, our approach involves storing the mean ($\mu \in \R^d$) and covariance ($\Sigma \in \R^{d\times d}$) of features. These stored features are then generated and replayed using Gaussian distribution to eliminate bias in the classifier, ensuring its proper alignment. Specifically, during the incremental training process, for $b$-th training stage, task-specific $\mathcal{A}_b$ is used to extract features from samples of all classes, calculating their mean and covariance:
\begin{equation}
    \begin{aligned}
        \mu_c=&\frac{1}{K}\sum\nolimits_{i=1}^{|\mathcal{D}^b|}\mathbb{I}(y_i=c)\phi(\x_i;\mathcal{A}_b),\\
        \Sigma_c=&\frac{1}{K}\sum\nolimits_{i=1}^{|\mathcal{D}^b|}\sum\nolimits_{j=1}^{|\mathcal{D}^b|}\mathbb{I}(y_i=c)\\
        &(\phi(\x_i;\mathcal{A}_b)-\mu_c)(\phi(\x_j;\mathcal{A}_b)-\mu_c),
    \end{aligned}
\end{equation}
where $K=\sum\nolimits_{i=1}^{|\mathcal{D}^b|}\mathbb{I}(y_i=c)$. This enables the model to mitigate the bias~\cite{zhao2020maintaining} related to the classifier after learning the task-specific adapter at each stage, thereby facilitating the alignment of the classifier.

Prior to each testing phase, the stored mean and covariance for each class are restored using Gaussian distribution. For each class $c \in \Y_b$, we generate features equivalent to five times the batchsize for aligning with the classifier:
\begin{equation}
        \hat{\phi}_c=\mathcal{N}(\mu_c,\Sigma_c),
\end{equation}
where $\hat{\phi}_c$ denotes a set of the generated features and $\mathcal{N}$ represents the Gaussian distribution. Since the progressively merged adapters we previously proposed can make all $\mathcal{A}_i$ have specificity while also having certain relevance, and because of the generalizability unique to the pre-trained models, the features extracted in this way and the generated features can align the classifiers well. Therefore, we can reduce the bias between classifiers using this method. This kind of bias is usually caused by the classifier being overconfident in new tasks and easily leads to catastrophic forgetting.

\section{Introduction about Compared Methods} \label{sec:3}
In this section, we present the details of the methods compared in the main paper. Each method utilizes the same pre-trained model (PTM) to ensure a fair comparison. These methods are enumerated as follows:
\begin{itemize}
    \item \textbf{Finetune}: updates all parameters with a PTM when continually trained on new tasks, but becomes susceptible to significant catastrophic forgetting. 
    \item \textbf{LwF}~\cite{li2017learning}: aims to resist forgetting by employing knowledge distillation~\cite{hinton2015distilling}, which creates a bridge between the last-stage model and the current one to transfer past knowledge.
    \item \textbf{L2P}~\cite{wang2022learning}: integrates visual prompt tuning~\cite{jia2022visual} into class-incremental learning using a pre-trained Vision Transformer~\cite{dosovitskiy2020image}. It further establishes a prompt pool, which facilitates the selection of instance-specific prompts.
    \item \textbf{DualPrompt}~\cite{wang2022dualprompt}: introduces two categories of prompts based on the L2P method: general prompts and expert prompts.
    \item \textbf{CODA-Prompt}~\cite{smith2023coda}: recognizes the limitations of instance-specific prompt selection. This approach seeks to overcome these challenges by prompt reweighting. Specifically, it improves the prompt selection process with an attention mechanism for prompt reweighting.
    \item \textbf{SimpleCIL}~\cite{zhou2024revisiting}: proproses a prototype-based classifier using PTM. Initializing with PTM, it establishes a prototype classifier for each category, employing a cosine classifier for the classification process.
    \item \textbf{APER}~\cite{zhou2024revisiting}: extends SimpleCIL by integrating both the pre-trained model and an adapted model. This approach considers the initial incremental stage as the sole adaptation phase, during which the PTM is tailored to extract task-specific features. Consequently, this model effectively unifies generalizability and adaptivity within a unified framework.
    \item \textbf{SLCA}~\cite{zhang2023slca}: extends the Gaussian modeling of previous classes in~\cite{Zhu_2021_CVPRb} to rectify classifiers during model updating.
    \item \textbf{EASE}~\cite{zhou2024expandable}: concatenates the feature representations of multiple task-specific backbones, leading to superior performance. It designs a semantic mapping strategy for classifier complement to compensate for the ever-expanding features and the previous classifiers.
\end{itemize}

\textbf{The methods described above are exemplar-free, meaning they do not necessitate the use of exemplars. On the other hand, we also evaluate several exemplar-based methods in the main paper as follows:}

\begin{itemize}
    \item \textbf{iCaRL}~\cite{rebuffi2017icarl}: employs knowledge distillation and exemplar-based replay to review previous knowledge. Additionally, it leverages the nearest center mean classifier for the final classification process.
    \item \textbf{DER}~\cite{yan2021dynamically}: utilizes a dynamically expandable representation to enhance incremental concept modeling more effectively.
    \item \textbf{FOSTER}~\cite{wang2022foster}: to reduce the memory burden associated with DER, this approach suggests the compression of backbones through knowledge distillation. Consequently, only a single backbone is maintained throughout the learning process. This method effectively enables feature expansion while minimizing memory consumption.
    \item \textbf{MEMO}~\cite{zhou2022model}: seeks to reduce the memory demands associated with DER from another strategy. It effectively segregates the network architecture into two distinct components: specialized (deep) layers and generalized (shallow) layers. This design enables the expansion of specialized layers while leveraging the existing generalized layers as a common foundation.
\end{itemize}

In the experiments, we reproduce the above methods based on their source code and PILOT\footnote{\url{https://github.com/sun-hailong/LAMDA-PILOT}}.

\section{More Results in Various Settings} \label{sec:4}
\begin{figure*}[t]
	\centering
	\begin{subfigure}{0.3\linewidth}
		\includegraphics[width=1\columnwidth]{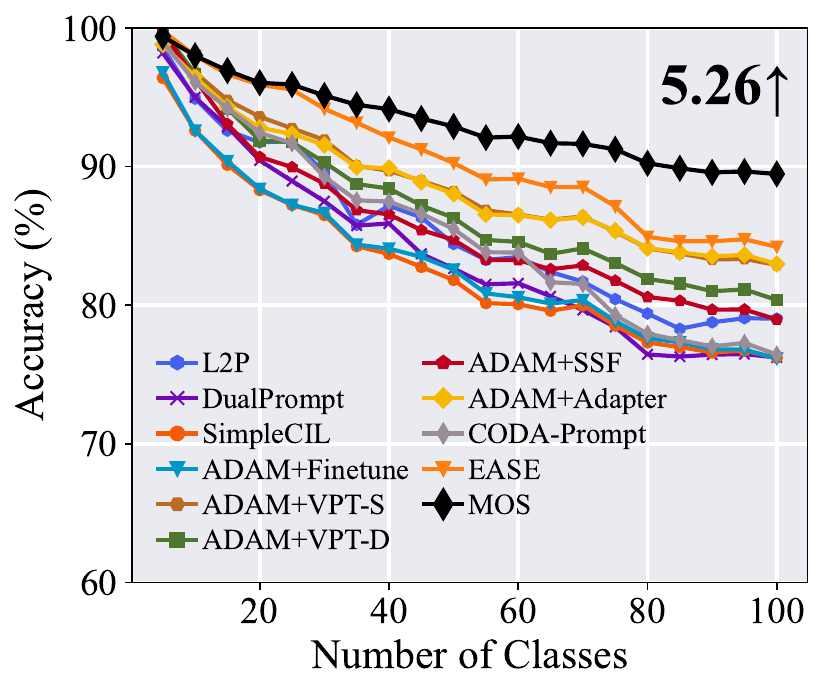}
		\caption{CIFAR100 B0 Inc5 IN1K}
		\label{fig:cifarb0inc5-in1k}
	\end{subfigure}
	\hfill
	\begin{subfigure}{0.3\linewidth}
		\includegraphics[width=1\linewidth]{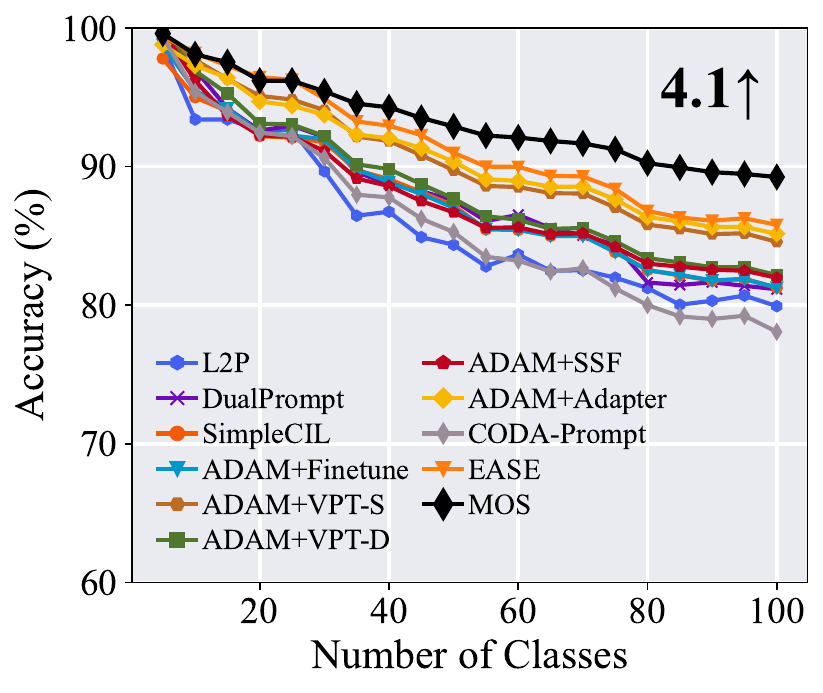}
		\caption{CIFAR100 B0 Inc5 IN21K}
		\label{fig:cifarb0inc5-in21k}
	\end{subfigure}
	\hfill
	\begin{subfigure}{0.3\linewidth}
		\includegraphics[width=1\linewidth]{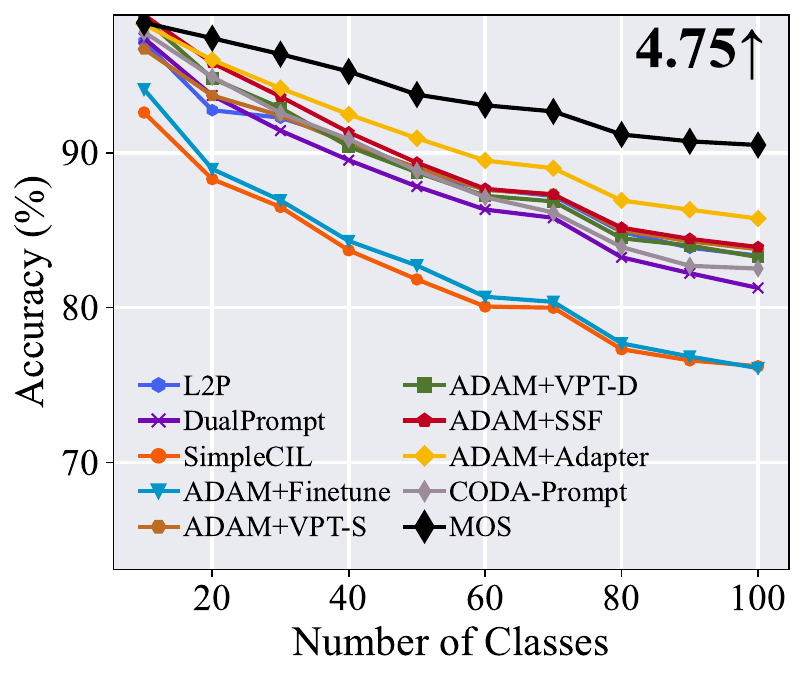}
		\caption{CIFAR100 B0 Inc10 IN1K}
		\label{fig:cifarb0inc10-in1k}
	\end{subfigure}
	\\
	\begin{subfigure}{0.3\linewidth}
		\includegraphics[width=1\linewidth]{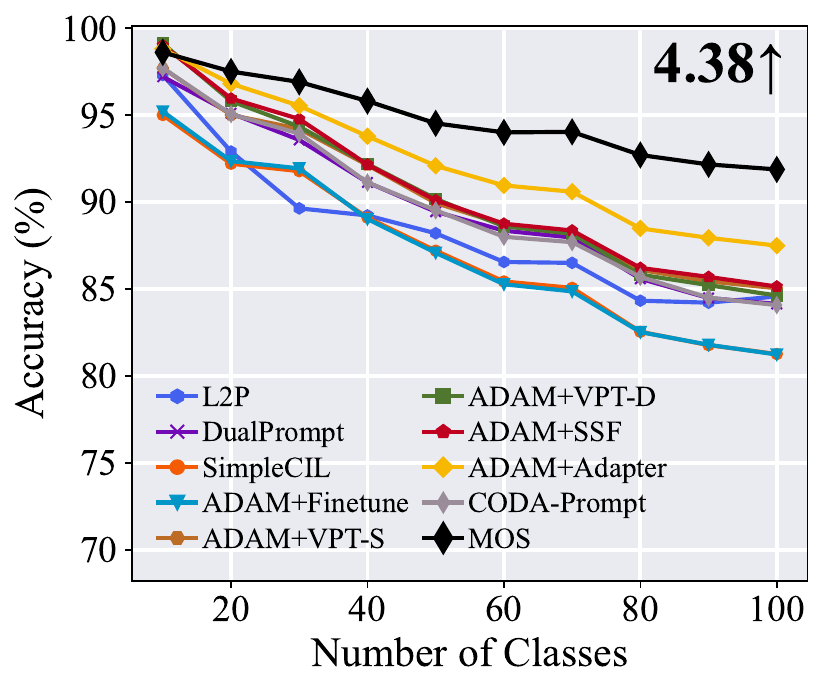}
		\caption{CIFAR100 B0 Inc10 IN21K}
		\label{fig:cifarb0inc10-in21k}
	\end{subfigure}
	\hfill
	\begin{subfigure}{0.3\linewidth}
		\includegraphics[width=1\linewidth]{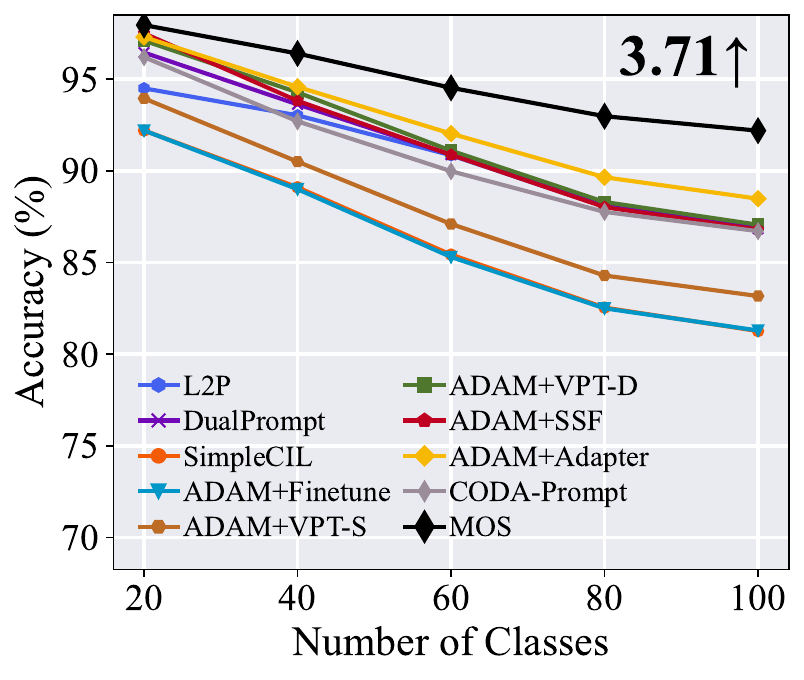}
		\caption{CIFAR100 B0 Inc20 IN21K}
		\label{fig:cifarb0inc20-in21k}
	\end{subfigure}
	\hfill
	\begin{subfigure}{0.3\linewidth}
		\includegraphics[width=1\columnwidth]{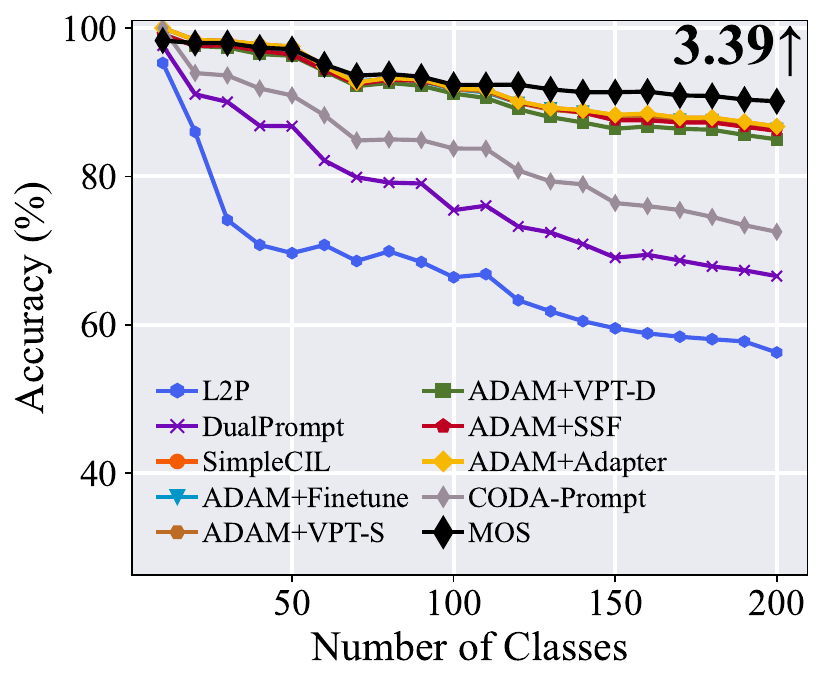}
		\caption{CUB200 B0 Inc10 IN21K}
		\label{fig:cubb0inc10-in21k}
	\end{subfigure}
 	\\
	\begin{subfigure}{0.3\linewidth}
		\includegraphics[width=1\linewidth]{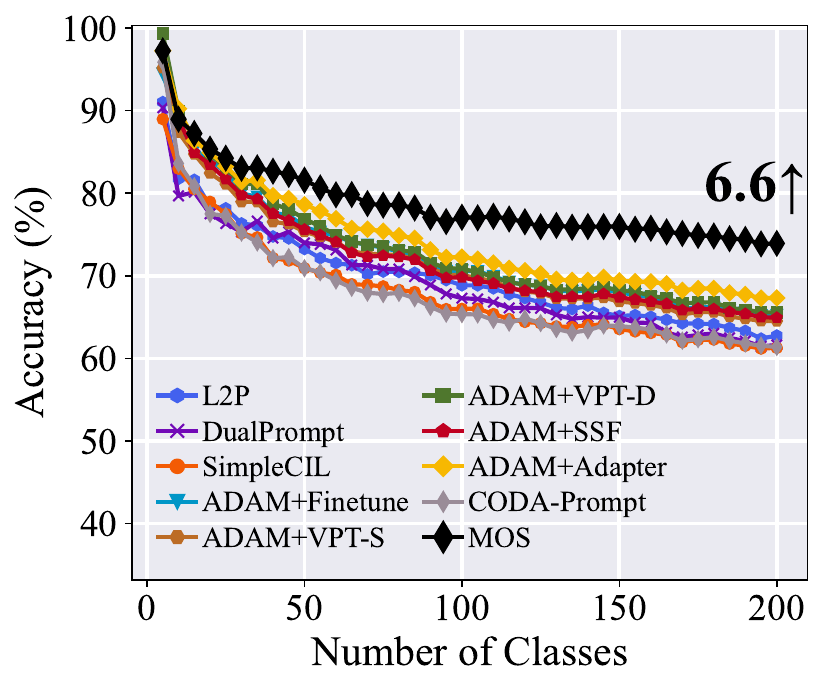}
		\caption{ImageNet-R B0 Inc5 IN1K}
		\label{fig:inrb0inc5-in1k}
	\end{subfigure}
	\hfill
	\begin{subfigure}{0.3\linewidth}
		\includegraphics[width=1\linewidth]{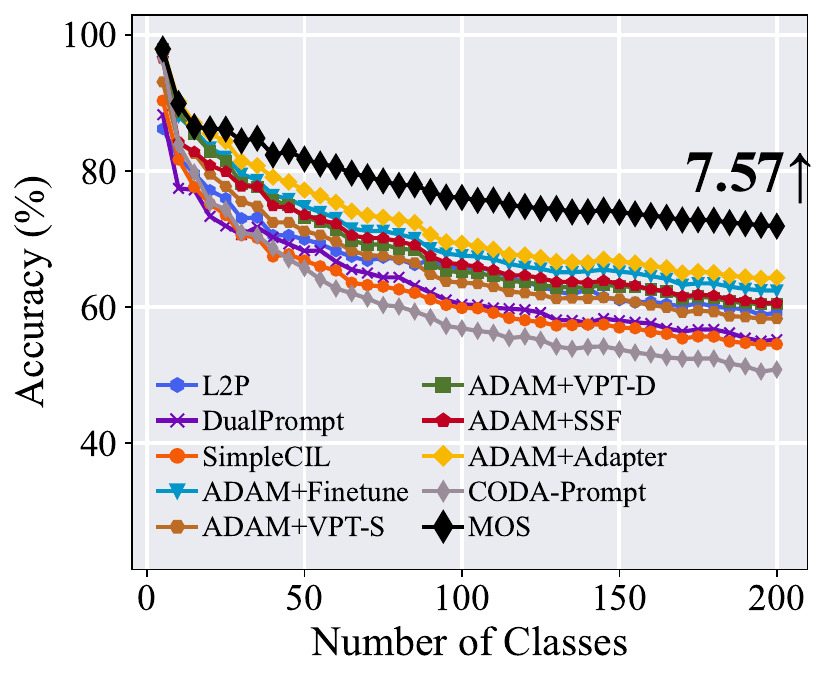}
		\caption{ImageNet-R B0 Inc5 IN21K}
		\label{fig:inrb0inc5-in21k}
	\end{subfigure}
	\hfill
	\begin{subfigure}{0.3\linewidth}
		\includegraphics[width=1\columnwidth]{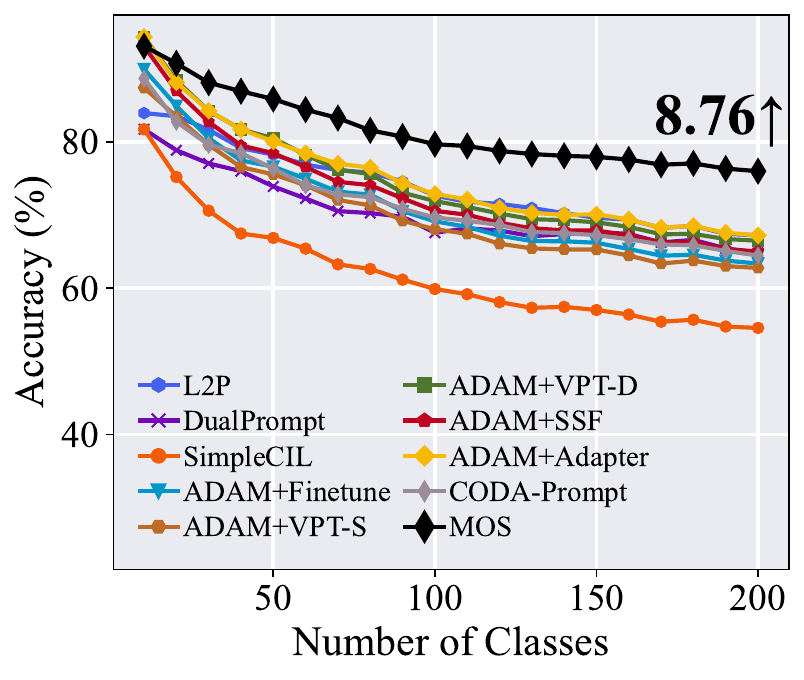}
		\caption{ImageNet-R B0 Inc10 IN21K}
		\label{fig:inrb0inc10-in21k}
	\end{subfigure}
  	\\
	\begin{subfigure}{0.3\linewidth}
		\includegraphics[width=1\linewidth]{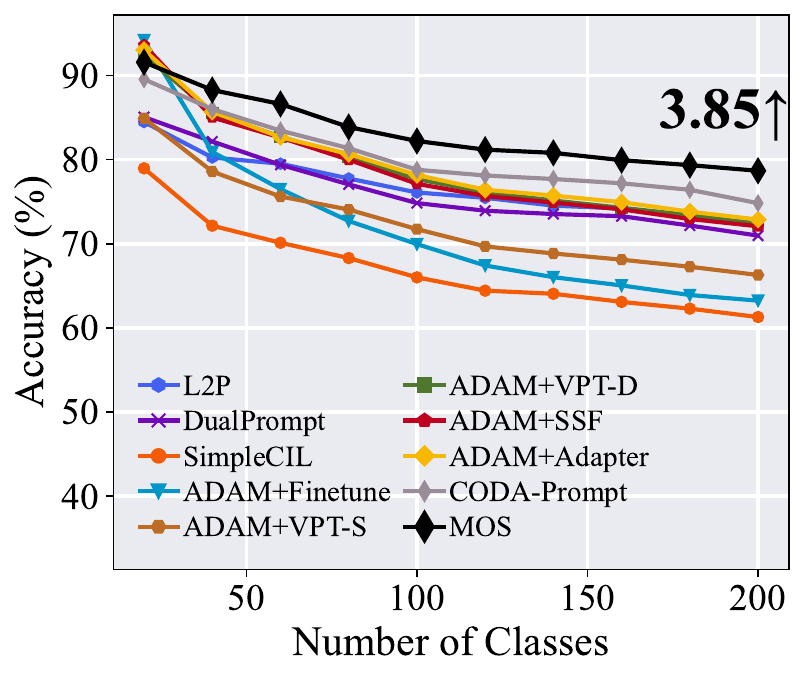}
		\caption{ImageNet-R B0 Inc20 IN1K}
		\label{fig:inrb0inc20-in1k}
	\end{subfigure}
	\hfill
	\begin{subfigure}{0.3\linewidth}
		\includegraphics[width=1\linewidth]{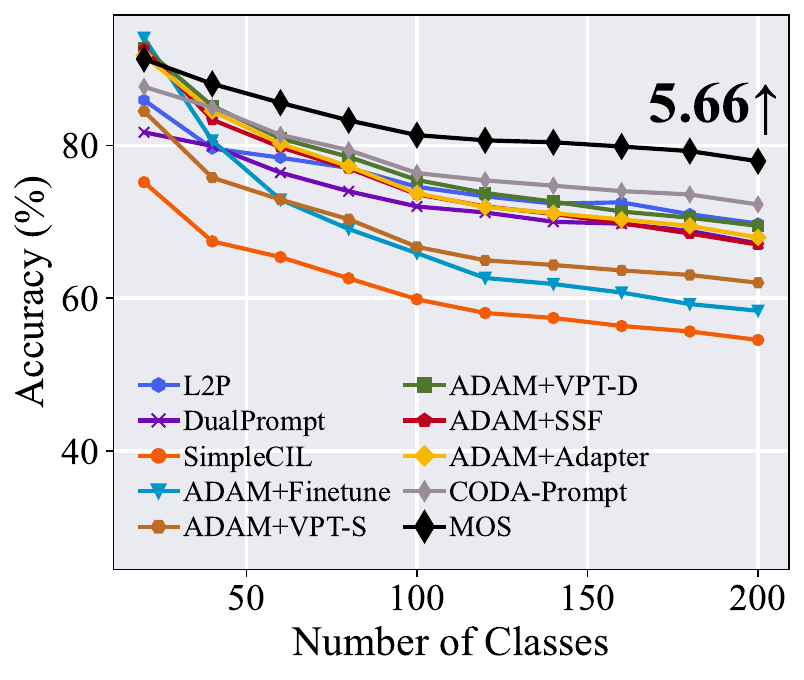}
		\caption{ImageNet-R B0 Inc20 IN21K}
		\label{fig:inrb0inc20-in2k}
	\end{subfigure}
	\hfill
	\begin{subfigure}{0.3\linewidth}
		\includegraphics[width=1\columnwidth]{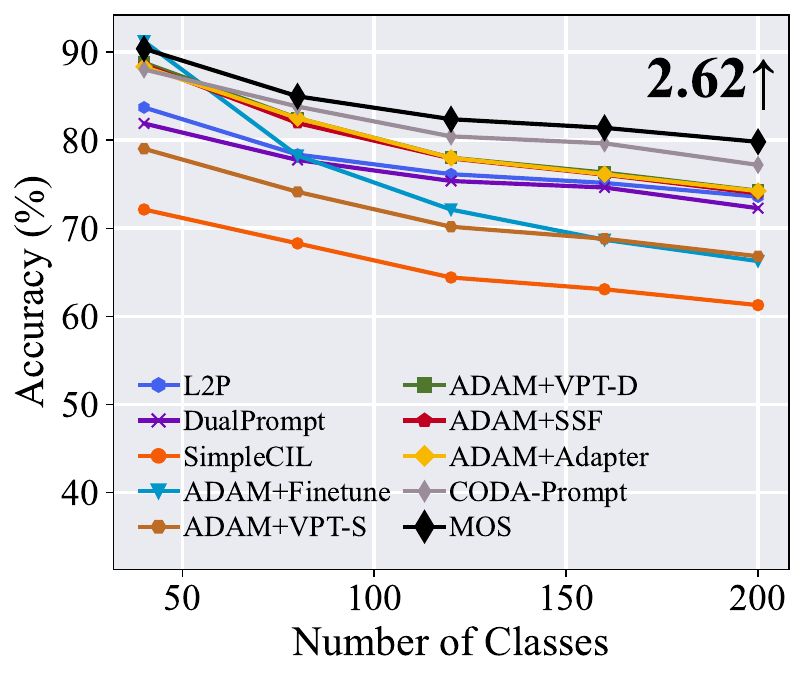}
		\caption{ImageNet-R B0 Inc40 IN1K}
		\label{fig:inrb0inc40-in1k}
	\end{subfigure}
	\caption{\small Performance curve of different methods under different settings. `IN21k' stands for {\bf ViT-B/16-IN21K} and `IN1K' stands for {\bf ViT-B/16-IN1K}. We annotate the relative improvement of \name above the runner-up method with numerical numbers at the last incremental stage. }
	\label{fig:benchmark1}
\end{figure*}

\begin{figure*}[t]
	\centering
	\begin{subfigure}{0.3\linewidth}
		\includegraphics[width=1\columnwidth]{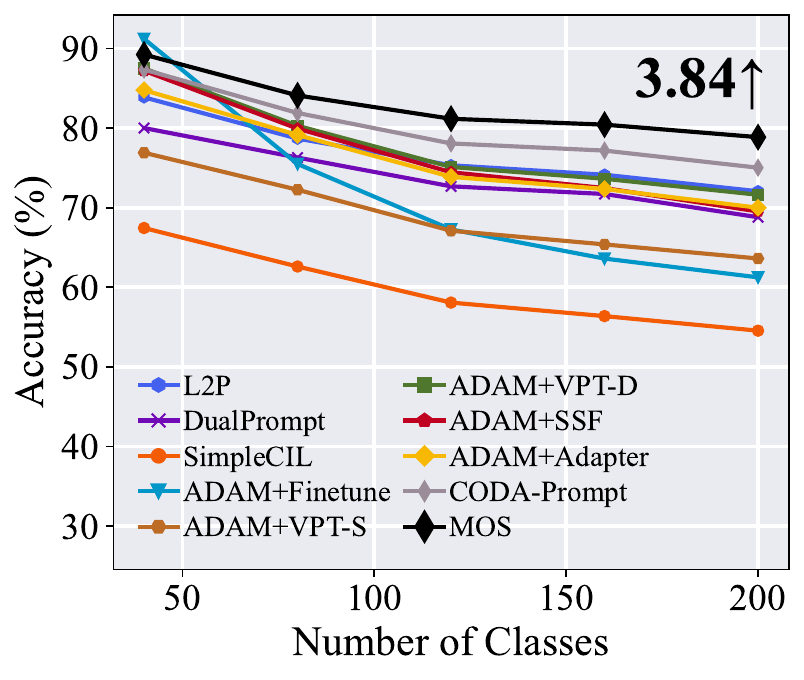}
		\caption{ImageNet-R B0 Inc40 IN21K}
		\label{fig:inrb0inc40-in21k}
	\end{subfigure}
	\hfill
	\begin{subfigure}{0.3\linewidth}
		\includegraphics[width=1\linewidth]{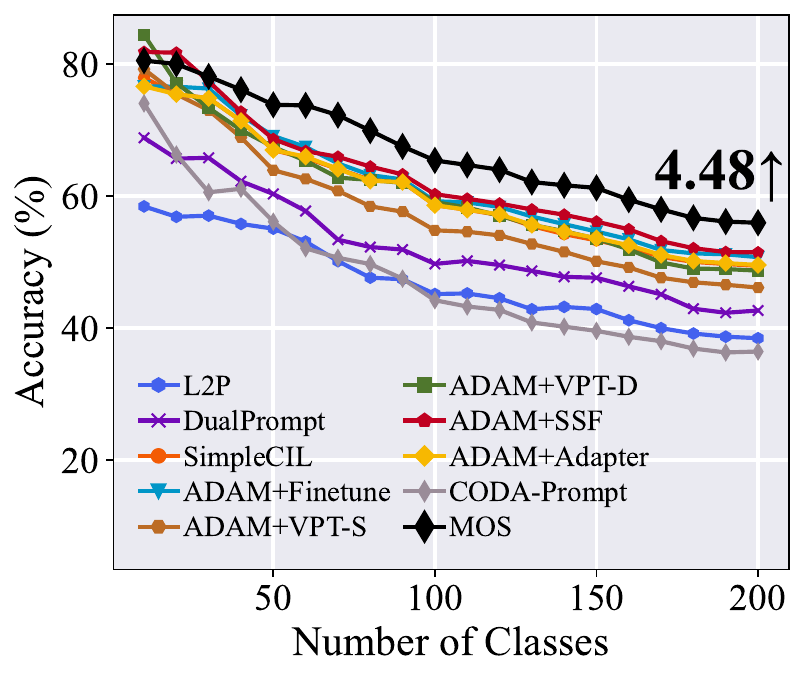}
		\caption{ImageNet-A B0 Inc10 IN1K}
		\label{fig:inab0inc10-in1k}
	\end{subfigure}
	\hfill
	\begin{subfigure}{0.3\linewidth}
		\includegraphics[width=1\linewidth]{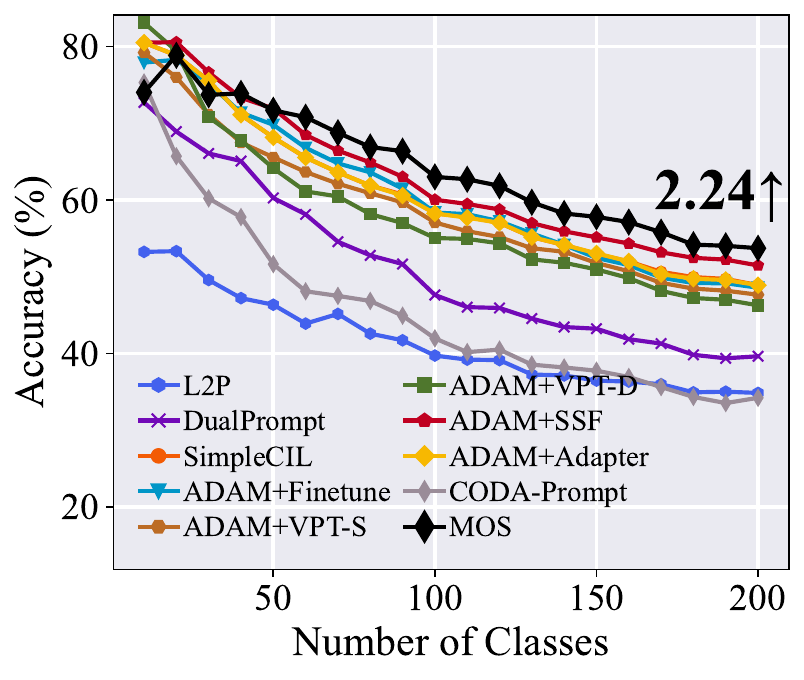}
		\caption{ImageNet-A B0 Inc10 IN21K}
		\label{fig:inab0inc10-in21k}
	\end{subfigure}
	\\
	\begin{subfigure}{0.3\linewidth}
		\includegraphics[width=1\linewidth]{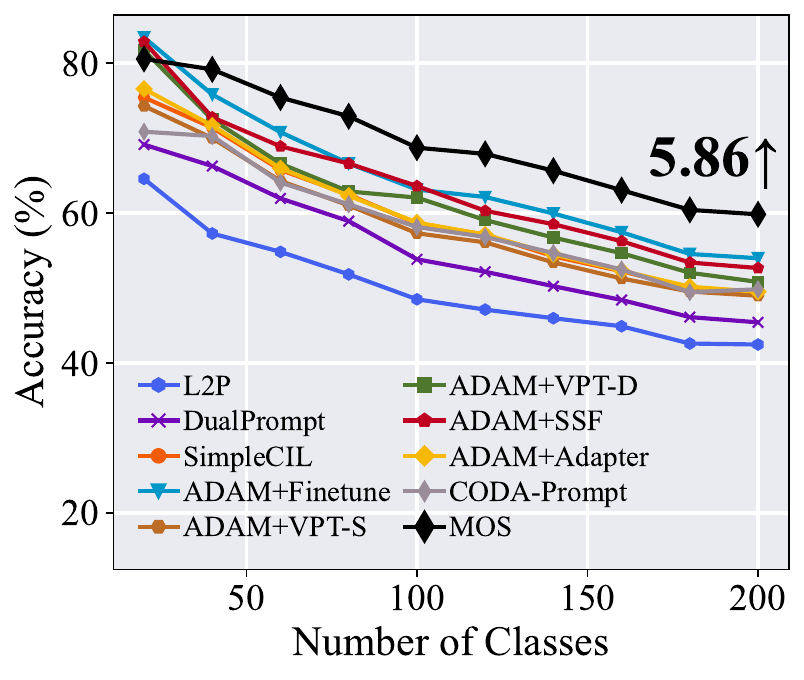}
		\caption{ImageNet-A B0 Inc20 IN1K}
		\label{fig:inab0inc20-in1k}
	\end{subfigure}
	\hfill
	\begin{subfigure}{0.3\linewidth}
		\includegraphics[width=1\linewidth]{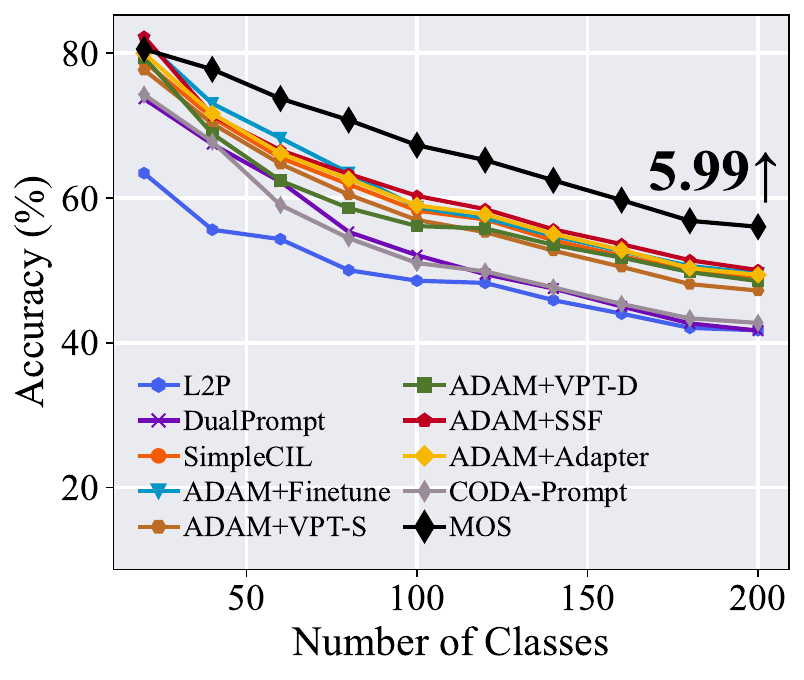}
		\caption{ImageNet-A B0 Inc20 IN21K}
		\label{fig:inab0inc20-in21k}
	\end{subfigure}
	\hfill
	\begin{subfigure}{0.3\linewidth}
		\includegraphics[width=1\columnwidth]{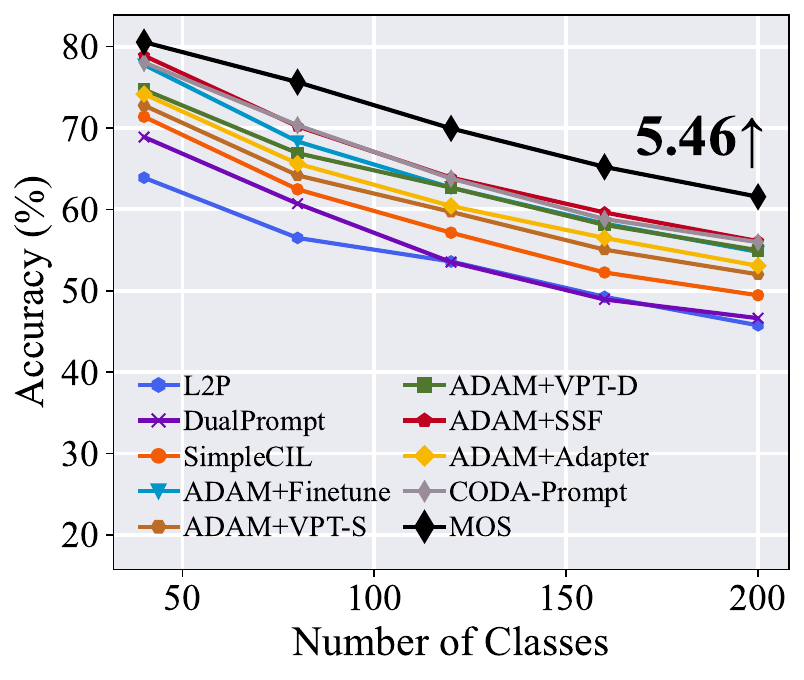}
		\caption{ImageNet-A B0 Inc40 IN1K}
		\label{fig:inab0inc40-in1k}
	\end{subfigure}
 	\\
	\begin{subfigure}{0.3\linewidth}
		\includegraphics[width=1\linewidth]{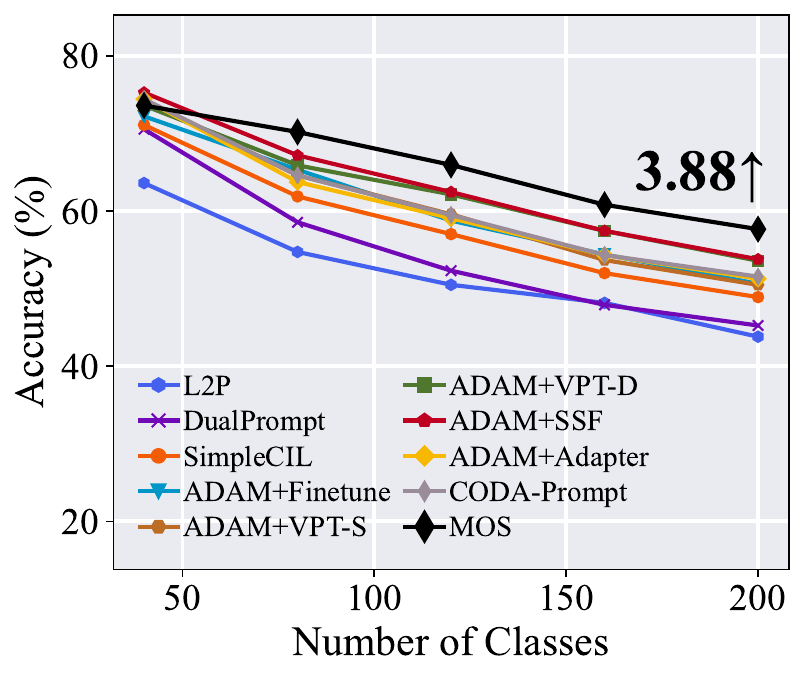}
		\caption{ImageNet-A B0 Inc40 IN21K}
		\label{fig:inab0inc40-in21k}
	\end{subfigure}
	\hfill
	\begin{subfigure}{0.3\linewidth}
		\includegraphics[width=1\linewidth]{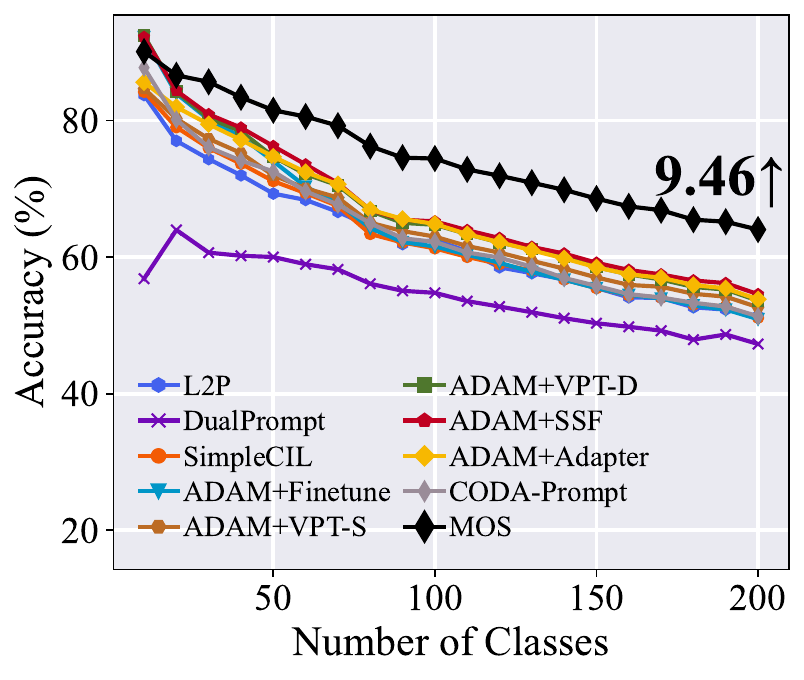}
		\caption{ObjectNet B0 Inc10 IN1K}
		\label{fig:objb0inc10-in1k}
	\end{subfigure}
	\hfill
	\begin{subfigure}{0.3\linewidth}
		\includegraphics[width=1\columnwidth]{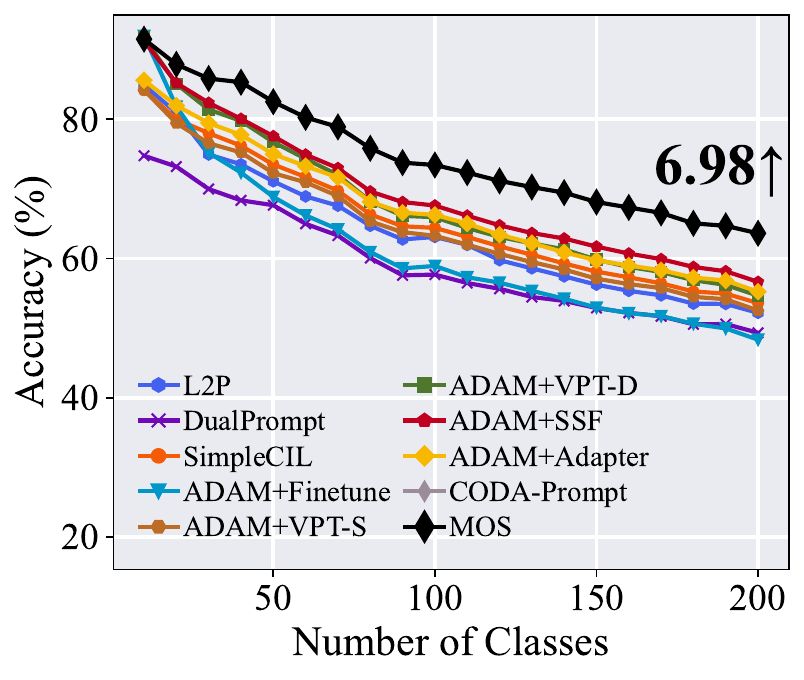}
		\caption{ObjectNet B0 Inc10 IN21K}
		\label{fig:objb0inc10-in21k}
	\end{subfigure}
  	\\
	\begin{subfigure}{0.3\linewidth}
		\includegraphics[width=1\linewidth]{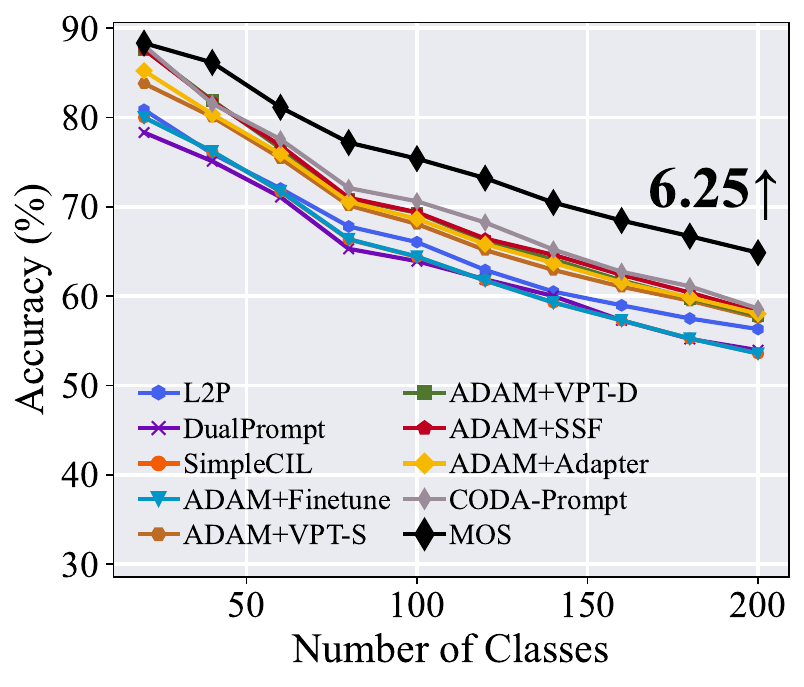}
		\caption{ObjectNet B0 Inc20 IN21K}
		\label{fig:objb0inc20-in21k}
	\end{subfigure}
	\hfill
	\begin{subfigure}{0.3\linewidth}
		\includegraphics[width=1\linewidth]{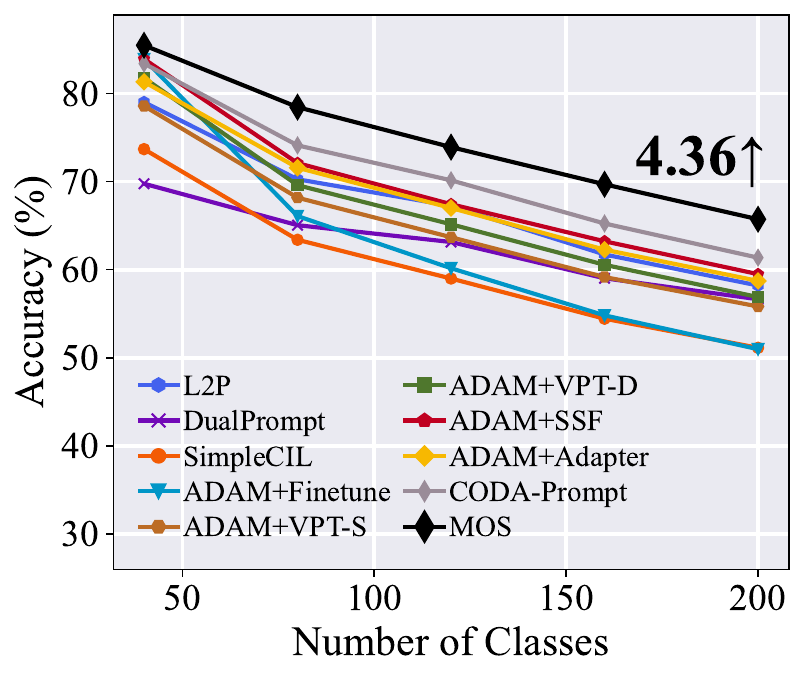}
		\caption{ObjectNet B0 Inc40 IN1K}
		\label{fig:objb0inc40-in1k}
	\end{subfigure}
	\hfill
	\begin{subfigure}{0.3\linewidth}
		\includegraphics[width=1\columnwidth]{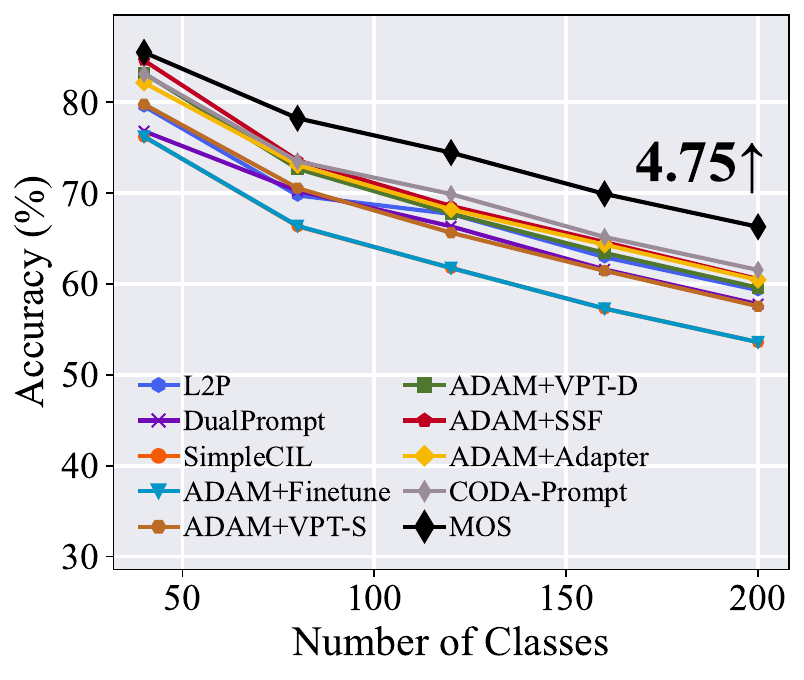}
		\caption{ObjectNet B0 Inc40 IN21K}
		\label{fig:objb0inc40-in21k}
	\end{subfigure}
	\caption{\small Performance curve of different methods under different settings. `IN21k' stands for {\bf ViT-B/16-IN21K} and `IN1K' stands for {\bf ViT-B/16-IN1K}. We annotate the relative improvement of \name above the runner-up method with numerical numbers at the last incremental stage. }
	\label{fig:benchmark2}
\end{figure*}

\begin{figure*}[t]
	\centering
	\begin{subfigure}{0.3\linewidth}
		\includegraphics[width=1\columnwidth]{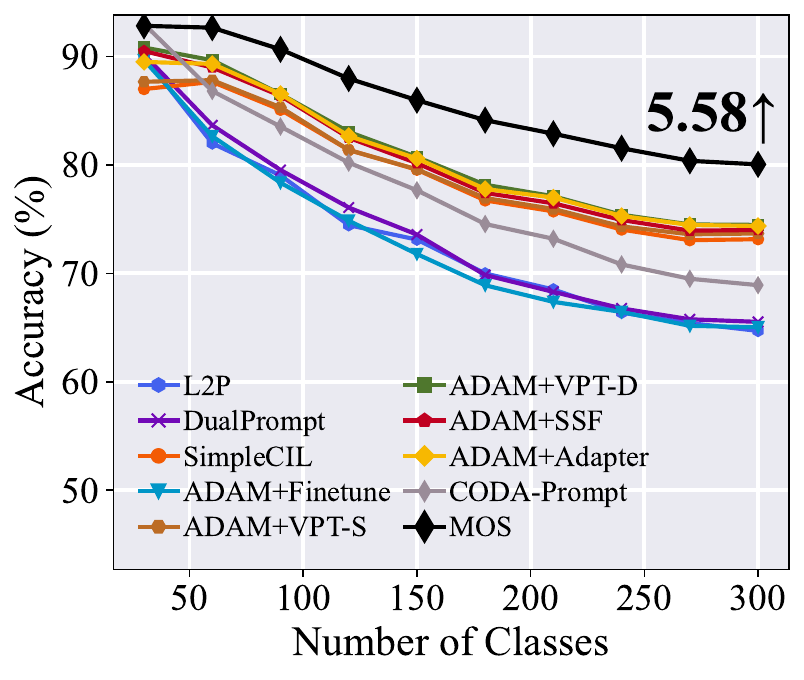}
		\caption{OmniBenchmark B0 Inc30 IN21K}
		\label{fig:omnib0inc30-in21k}
	\end{subfigure}
	\hfill
	\begin{subfigure}{0.3\linewidth}
		\includegraphics[width=1\linewidth]{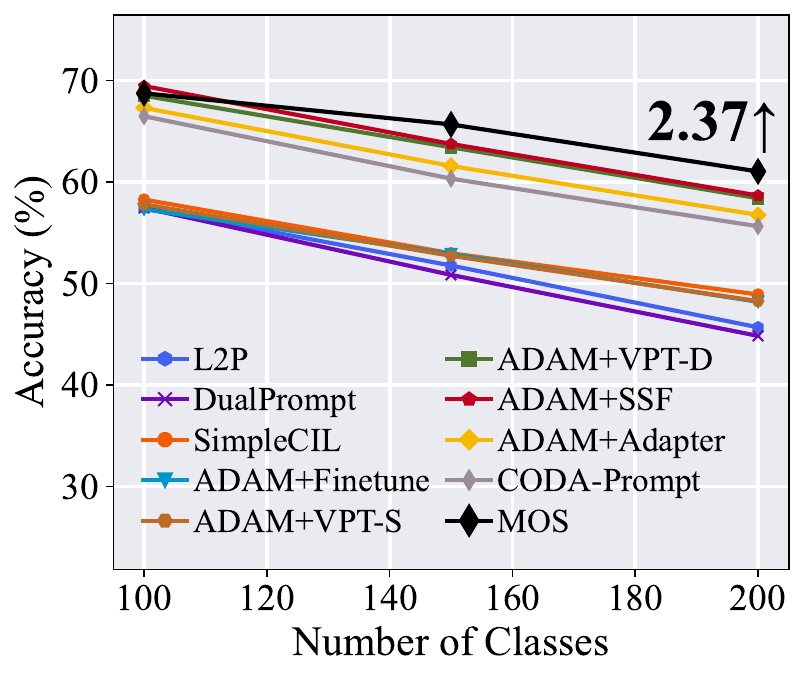}
		\caption{ImageNet-A B100 Inc50 IN21K}
		\label{fig:inab100inc50-in1k}
	\end{subfigure}
	\hfill
	\begin{subfigure}{0.3\linewidth}
		\includegraphics[width=1\linewidth]{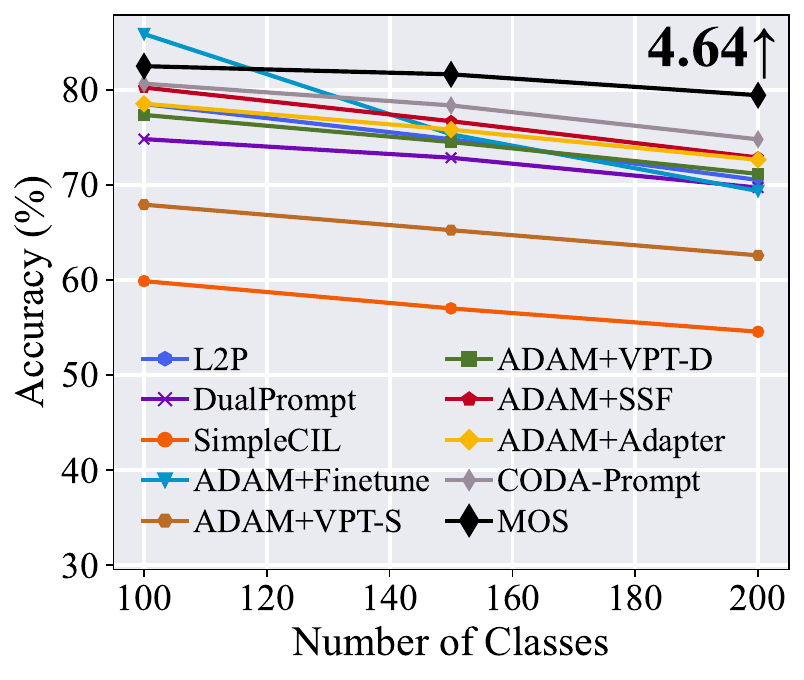}
		\caption{ImageNet-R B100 Inc50 IN21K}
		\label{fig:inrb100inc50-in1k}
	\end{subfigure}
	\caption{\small Performance curve of different methods under different settings. `IN21k' stands for {\bf ViT-B/16-IN21K} and `IN1K' stands for {\bf ViT-B/16-IN1K}. We annotate the relative improvement of \name above the runner-up method with numerical numbers at the last incremental stage. }
	\label{fig:benchmark3}
\end{figure*}

In this section, we present more experimental results from various methods. We specifically detail the incremental performance of these methods using ViT-B/16-IN21K and ViT-B/16-IN1K, as depicted in Figure~\ref{fig:benchmark1},  Figure~\ref{fig:benchmark2}, and Figure~\ref{fig:benchmark3}. The results demonstrate that \name consistently surpasses other methods across different datasets, achieving a significant margin of superiority.

\end{document}